%% file: main-arxiv.tex
\documentclass{article} 
\usepackage{iclr2026_conference,times}

\input{math_commands.tex}

\iclrfinalcopy

\makeatletter
\def\ps@headings{%
\def\@oddhead{\hfill \hfill} 
\def\@evenhead{\hfill \hfill}
\def\@oddfoot{\hfill\thepage\hfill}
\def\@evenfoot{\hfill\thepage\hfill}}
\makeatother
\usepackage{graphicx}
\usepackage{booktabs}
\usepackage{amsmath}
\usepackage{amssymb}
\usepackage{algorithm}
\usepackage{algorithmic}
\usepackage{subcaption}
\usepackage{multirow}
\usepackage{wrapfig}
\usepackage{float}
\usepackage{xcolor}
\usepackage{hyperref}
\usepackage{url}

\hypersetup{
  colorlinks=true,
  linkcolor=blue,
  citecolor=blue,
  urlcolor=blue
}

\newcommand{\std}[1]{{\scriptsize $\pm$#1}}
\newcommand{\best}[1]{\textbf{#1}}
\newcommand{\second}[1]{\underline{#1}}

\title{InstantForget: Update-Free Backdoor Unlearning with Inference-Time Feature Reset}

\author{Zhenyu Yu \\
College of Computer Science and Artificial Intelligence\\
Fudan University \\
\texttt{yuzhenyuyxl@foxmail.com}
}

\begin{document}

\maketitle

\begin{abstract}
Backdoor unlearning aims to remove a malicious trigger behavior from a deployed model while preserving clean utility. We study the update-free inference-time setting, where model parameters remain frozen. First, we audit a common projection assumption under oracle paired clean and triggered features. Projection succeeds mainly on BadNets and leaves WaNet, Blended, and SIG at 0.683, 0.888, and 0.941 ASR on CIFAR-10 ResNet-18. This failure is not explained by spectral compactness, spatial locality, or subspace misalignment. It is predicted by a logit-triplet gap involving the target margin, target-logit drop, and non-target logit rise. We then introduce \textbf{InstantForget}, a clean-calibrated gated reset that flags anomalous features with a Mahalanobis score and moves only flagged features toward a neutral non-target representation. With one fixed operating point selected on held-out triggered validation, InstantForget reduces average ASR to 0.071 across four non-adaptive CIFAR-10 triggers without triggered samples or parameter updates at deployment. It also reaches 0.981 detection AUROC and transfers to six of eight tested backbones. Reported failures under WaNet, ModelNet10 point blend, two backbone geometries, and adaptive feature-compactness attacks define the method's scope.
\end{abstract}


\section{Introduction}
\label{sec:intro}

Backdoor attacks train a model to behave normally on clean inputs while mapping triggered inputs to an attacker-chosen target. They have been demonstrated with patch triggers, blended triggers, sinusoidal triggers, sample-specific triggers, and geometric warps~\citep{gu2019badnets,chen2017targetedbackdoorattacksdeep,barni2019new,li2021invisible,nguyen2021wanet}. A large defense literature detects or repairs such models through activation statistics, trigger inversion, input perturbation, pruning, distillation, adversarial unlearning, and benchmarked post-training pipelines~\citep{tran2018spectral,wang2019neural,gao2019strip,liu2018fine,li2021neural,zeng2022adversarial,wu2022backdoorbench,li2023rnp,su2025burnbackdoorunlearningadversarial,abad2025soklinedefensebackdoor}. These defenses are measured primarily by attack success rate (ASR), the fraction of triggered inputs still classified as the target, yet a low ASR alone can mask a degenerate fix that suppresses the trigger by destroying the target class itself.

Backdoor unlearning asks for a more selective outcome. The malicious trigger-to-target association should be forgotten while clean accuracy and target-class utility are preserved. This differs from certified data deletion and from standard machine unlearning, where the target is usually the influence of training examples rather than a deployed trigger behavior~\citep{bourtoule2021machine,guo2020certified,neel2021descent,koloskova2025icml-certified,cooper2025neurips-machine}. The setting is especially constrained after deployment, where retraining data, optimizer state, or permission to update parameters may be unavailable. We therefore study \emph{update-free inference-time backdoor unlearning}. The defender may edit representations of a frozen model at inference time, but may not modify the model parameters.

A natural update-free unlearning strategy is projection. Given paired clean and triggered features, the defender estimates a trigger-induced subspace $D=H_t-H_c$ and suppresses it before the classifier head. Related representation-editing methods remove linearly encoded concepts or forget-specific directions through closed-form feature or head edits~\citep{belrose2023neurips-leace,hatami2026classunlearningdepthawareremoval}. Projection is simple, cheap, and compatible with frozen models, and it assumes that the trigger behavior is concentrated in removable feature directions. We test this assumption under favorable conditions and find a sharp trigger-dependent failure that is diagnostic rather than accidental. It is not explained by poor access, insufficient hyperparameter search, or a missing subspace direction. Spatial localization explains why a patch trigger is different but not the graded difficulty among distributed triggers, spectral compactness is unstable across seeds, and principal-angle diagnostics show that SVD identifies a target-logit pathway for all triggers. The reliable predictor is downstream of the edit: a triggered prediction flips only when the target margin is smaller than the target-logit drop plus the non-target-logit rise. This logit-triplet gap gives a compact explanation for when projection changes the decision and when it merely suppresses a target-aligned feature without unlearning the trigger behavior.

This diagnosis motivates InstantForget. Projection is useful for auditing, but it is a poor unlearning gate because its direction is nearly collinear with the target-class head and therefore confounds triggered inputs with clean target-class features. InstantForget decouples detection from correction. It fits a clean-calibrated Mahalanobis gate in feature space, calibrates the clean false-positive rate, and resets only flagged features toward a neutral non-target representation. The method is deliberately scoped. It targets non-adaptive backdoors in a frozen model, uses clean calibration data at deployment, and treats target-class utility as a first-class outcome rather than a secondary accuracy statistic. We evaluate it against post-training repair defenses, closed-form erasure editors, and inference-time input detectors under one protocol. We then probe its boundaries with cross-domain datasets, eight backbone architectures, alternative target labels, and an adaptive feature-compactness attacker.

Our \textbf{contributions} are:
\begin{itemize}
\item \textbf{Projection audit for update-free backdoor unlearning.} We evaluate an oracle SVD-of-$D$ projection family over four triggers, and a broad editor sweep, showing that projection-based unlearning has a trigger-dependent failure gap even under favorable conditions.
\item \textbf{Mechanistic diagnostic for projection-based unlearning.} We derive and validate the logit-triplet gap $m-(\delta_t+\delta_o)$, which predicts whether projection will flip triggered predictions away from the target, unlike spectral compactness or subspace-alignment diagnostics. The diagnosis extends to closed-form concept erasure. LEACE and a DAMP-style prototype-residual head edit both receive oracle paired trigger features, yet leave ASR essentially unchanged in our benchmark.
\item \textbf{InstantForget, an update-free gated reset.} InstantForget decouples detection from correction with a Mahalanobis feature gate and neutral reset, using only clean data and no parameter updates at deployment. We compare it against four standard BackdoorBench post-training defenses, closed-form erasure editors, and the input-level detectors STRIP and SCALE-UP. We also sweep operating points and stress-test target labels, cross-domain datasets, eight backbone architectures, and adaptive feature-compactness attacks.
\end{itemize}

\section{Related Work}
\label{sec:related}

\paragraph{Backdoor attacks and defenses.} Backdoor attacks train a model to retain high clean accuracy while mapping triggered inputs to an attacker-chosen target. BadNets established the canonical patch-trigger setting~\citep{gu2019badnets}. Later attacks reduce visibility or locality through sample-specific triggers~\citep{li2021invisible}, warping~\citep{nguyen2021wanet}, and other distributed patterns. Post-training defenses modify the trained model through pruning, distillation, adversarial unlearning, or cleansing ~\citep{liu2018fine,wang2019neural,gao2019strip,li2021neural,li2021abl,wu2021adversarial,zeng2022adversarial,pang2023backdoor,li2023rnp}. Recent work continues to improve backdoor unlearning and evaluation, including boundary-based unlearning, contamination-robust fine-tuning, diffusion-model backdoor unlearning, and systematic analyses of defense evaluation ~\citep{wu2022backdoorbench,zhu2024neurips-breaking,su2025burnbackdoorunlearningadversarial,wei2023shared,jiang2025backdoortokenunlearningexposing,abad2025soklinedefensebackdoor}. InstantForget differs in access regime. It freezes the model, uses no triggered examples during deployment calibration, and reports target-class clean accuracy so that ASR reduction cannot be mistaken for destroying the target class.
\vspace{-10pt}

\paragraph{Machine unlearning and its limits.} Machine unlearning studies how to remove the effect of training data from a model ~\citep{bourtoule2021machine,neel2021descent,guo2020certified,xu2023survey}. Newer work has expanded both the theory and the warning signs: certified neural-network unlearning, in- and out-of-distribution unlearning trade-offs, targeted single-layer unlearning, and position papers all emphasize that unlearning claims must be scoped and evaluated carefully~\citep{koloskova2025icml-certified,allouah2025iclr-utility,cai2025targeted,cooper2025neurips-machine}, a concern echoed by benchmarks for selective forgetting in generative models~\citep{yu2025forgetme}. Most relevant to our setting, recent studies show that unlearning can fail to remove poisoning effects or can itself become a security surface ~\citep{pawelczyk2025iclr-machine,lu2025badfubackdoorfederatedlearning,alam2025reveilunconstrainedconcealedbackdoor}. InstantForget is not certified data deletion. It studies behavioral unlearning of a specific trigger-to- target mapping in a frozen classifier.
\vspace{-14pt}

\paragraph{Representation editing and clean-calibrated gates.} Projection-based mitigation and representation editing estimate a subspace associated with an unwanted behavior and suppress it at inference time. Closely related defenses use representation statistics, entropy perturbations, trigger inversion, or neuron selection to detect or remove backdoor behavior ~\citep{tran2018spectral,chen2018detectingbackdoorattacksdeep,gao2019strip,wang2019neural,li2023rnp,zhu2024neurips-breaking}. Input-level detection methods such as STRIP~\citep{gao2019strip} and SCALE-UP~\citep{guo2023scaleup} flag triggered inputs at inference time without modifying the model. We compare against both under a matched calibration protocol. Closed-form concept erasure is an adjacent family. LEACE~\citep{belrose2023neurips-leace} removes a linearly encoded concept with a minimal affine edit, and concurrent work DAMP~\citep{hatami2026classunlearningdepthawareremoval} performs one-shot projection surgery on layer weights to unlearn classes, observing, as we do, that output-level forgetting can mask intact internal representations. Our setting differs in task and conclusion. The concept to remove is a backdoor trigger whose feature direction is nearly collinear with the clean target class, and we show empirically that erasing that direction, even with oracle trigger access, does not unlearn the trigger behavior. Feature unlearning has also been studied in federated settings, where methods such as Ferrari optimize feature sensitivity under explicit unlearning requests. Subspace projection is attractive because it is cheap and compatible with frozen models, but it is not automatically selective unlearning. InstantForget contributes a negative and constructive result for this family. Oracle projection can identify target-aligned directions without providing a safe correction rule. The final method separates detection from correction. A clean-calibrated Mahalanobis gate controls false positives, and a neutral non-target reset specifies how flagged features are changed rather than merely rejecting suspicious inputs.

\begin{figure}
    \centering
    \includegraphics[width=1.0\linewidth]{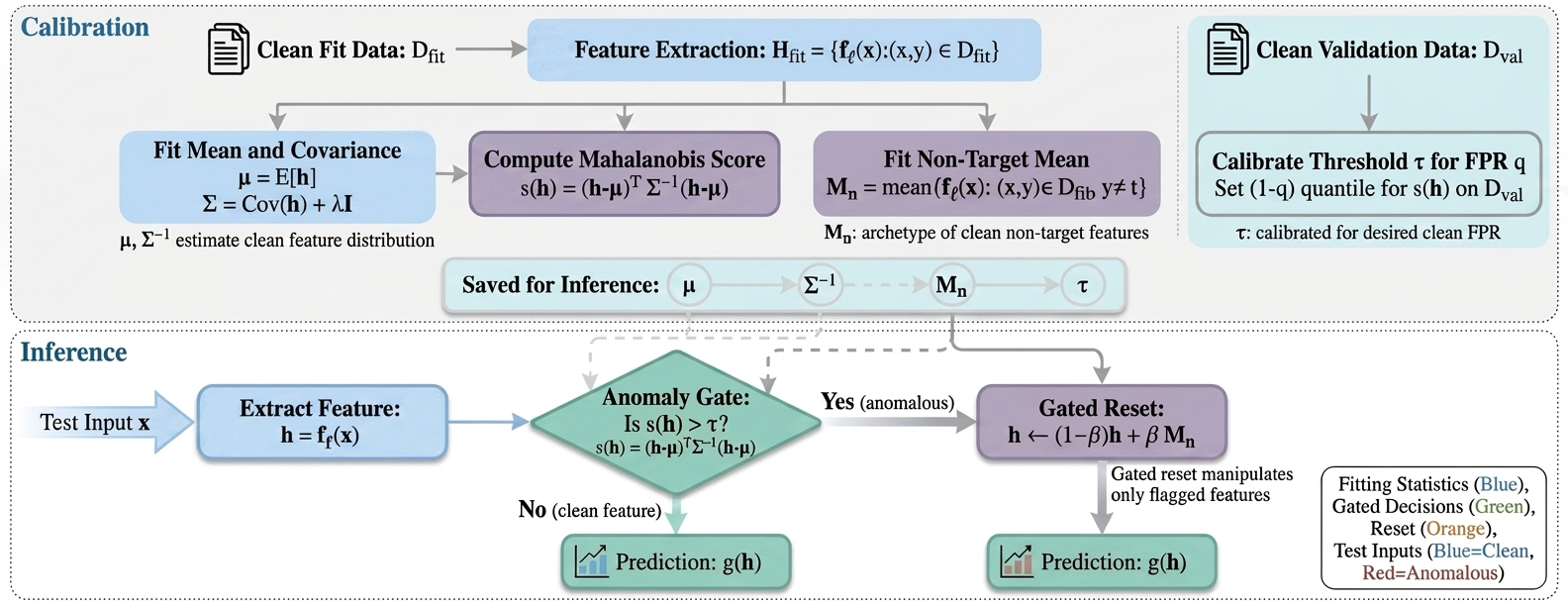}
    \caption{Overview of InstantForget. \textbf{Calibration} (clean data only): from a clean fit split $D_{\mathrm{fit}}$ we estimate the clean feature mean $\mu$ and covariance $\Sigma$ for the Mahalanobis score $s(h)=(h-\mu)^\top\Sigma^{-1}(h-\mu)$, store the non-target mean $M_n$ as a neutral reset target, and set the threshold $\tau$ to the $(1-q)$ quantile of $s$ on a clean validation split $D_{\mathrm{val}}$ (clean FPR $q$). \textbf{Inference}: a test feature $h=f_\ell(x)$ is passed unchanged to the frozen head $g$ when $s(h)\le\tau$, and otherwise gated-reset by $h\leftarrow(1-\beta)h+\beta M_n$ toward the neutral non-target mean. The reset touches only flagged features and updates no model parameters. Only $(\mu,\Sigma^{-1},M_n,\tau)$ are saved; no triggered data is used.}
    \label{fig:overview}
\end{figure}
\vspace{-10pt}

\section{Method}
\label{sec:method}

\subsection{Access Regime and Evaluation Goal}
\label{sec:background}

Let $f_\theta=g\circ f_\ell$ be a frozen classifier, where $h=f_\ell(x)\in\mathbb{R}^d$ is the feature at layer $\ell$ and $g(h)=Wh+b$ is the linear classification head. A backdoor trigger $T_\eta$ maps an input $x$ to a triggered input $T_\eta(x)$, which the compromised model predicts as a target class $t$. We evaluate unlearning using attack success rate (ASR) on triggered non-target inputs, clean accuracy, and target-class clean accuracy. The last metric is essential: a degenerate unlearning method can drive ASR to zero by destroying the target class.

We focus on update-free inference-time editing. An unlearning method may use forward passes through the frozen model and clean calibration data, but it does not update model parameters. We consider three access regimes. The audit in Sec.~\ref{sec:audit} grants projection methods paired clean/triggered features to measure their best possible behavior. InstantForget in Sec.~\ref{sec:instantforget} is stricter at deployment. It estimates its gate and reset statistics from clean data only and requires no trigger pattern. The fixed operating point used in the main tables is selected once on held-out triggered validation (Sec.~\ref{sec:experiments}). We treat that as design-time model selection, not deployment calibration. Throughout the paper, \emph{clean-calibrated} refers to the deployed estimator and threshold, while \emph{clean-only at deployment} means no triggered examples are used to fit $(\mu,\Sigma,M_n,\tau)$ on the deployed model.

\subsection{Oracle Projection Audit}
\label{sec:projection}

Given paired clean and triggered features $H_c,H_t\in\mathbb{R}^{n\times d}$, define
\begin{equation}
  D = H_t - H_c .
\end{equation}
The SVD-of-$D$ editor computes the top-$k$ right singular vectors $U\in\mathbb{R}^{d\times k}$ and applies the shrinkage projection
\begin{equation}
  P_\alpha = I - \alpha U U^\top,\qquad \alpha\in(0,1].
\end{equation}
The edited prediction is $g(P_\alpha h)$. This family covers the core operation used by projection-style update-free editors: estimate a trigger direction, then suppress it. We do not claim that every update-free method is exactly this operator. Rather, we isolate the projection core under oracle paired-feature access so that failure reflects edit geometry rather than subspace-estimation noise.

\paragraph{Evaluation objective.} For an unlearned model, ASR alone is insufficient: a method can obtain low ASR by destroying clean accuracy, or more subtly by destroying the target class. In the projection audit we therefore select operating points by
\begin{equation}
  J_\kappa = \mathrm{ASR}+\kappa\max(\Delta\mathrm{Acc},0),
\label{eq:jscore}
\end{equation}
where $\Delta\mathrm{Acc}$ is the clean-accuracy drop relative to the undefended backdoored model. The main text uses $\kappa=1$ and reports ASR at the selected operating point. Supplementary sweeps with $\kappa\in\{0.5,1,2\}$ preserve the same trigger ranking. For the final clean-calibrated defense, we additionally report target-class clean accuracy and the realized clean false-positive rate.

\subsection{Why Projection Fails at the Logit Level}
\label{sec:logit_condition}

Projection removes a feature component, but unlearning is evaluated after the linear head. The relevant question is therefore not whether a trigger direction is found, but whether the induced logit change flips the triggered prediction away from the target. For a triggered feature $h_t$ with target class $t$, define the original target margin
\begin{equation}
  m(h_t)=\operatorname{logit}_t(h_t)
  -\max_{c\neq t}\operatorname{logit}_c(h_t).
\label{eq:target_margin}
\end{equation}
For a rank-1 projection $P_1=I-u_1u_1^\top$, define the target-logit drop
\begin{equation}
  \delta_t(h_t)
  =\operatorname{logit}_t(h_t)-\operatorname{logit}_t(P_1h_t)
  =(h_t^\top u_1)(u_1^\top W_t),
\label{eq:target_drop}
\end{equation}
and the best non-target logit rise
\begin{equation}
  \delta_o(h_t)
  =\max_{c\neq t}\operatorname{logit}_c(P_1h_t)
  -\max_{c\neq t}\operatorname{logit}_c(h_t).
\label{eq:nontarget_rise}
\end{equation}

\paragraph{Proposition 1 (projection flip condition).} If $h_t$ is classified as the target before projection, then the projected feature $P_1h_t$ is classified as a non-target class whenever
\begin{equation}
  m(h_t)-\bigl(\delta_t(h_t)+\delta_o(h_t)\bigr)<0.
\label{eq:flip_condition}
\end{equation}
Conversely, if the left-hand side is positive and the maximizer among non-target classes is unchanged after projection, the prediction remains the target.

\paragraph{Proof.} Let $o=\arg\max_{c\neq t}\operatorname{logit}_c(h_t)$ and $o'=\arg\max_{c\neq t}\operatorname{logit}_c(P_1h_t)$. The target prediction is removed exactly when
\begin{equation}
  \operatorname{logit}_{o'}(P_1h_t)>\operatorname{logit}_t(P_1h_t).
\label{eq:flip_proof_start}
\end{equation}
Substituting the definitions above gives
\begin{equation}
  \max_{c\neq t}\operatorname{logit}_c(h_t)+\delta_o(h_t)
  >
  \operatorname{logit}_t(h_t)-\delta_t(h_t),
\label{eq:flip_proof_substitute}
\end{equation}
which is equivalent to Eq.~(\ref{eq:flip_condition}). If the non-target maximizer is unchanged and the inequality is reversed, the target logit remains larger than the best non-target logit after projection. \hfill $\square$

This condition explains why spectral compactness alone is insufficient. Even if a low-rank direction captures the trigger-induced shift, projection succeeds only when the target margin is overcome by the combined decrease of the target logit and increase of the strongest non-target logit. The experiments therefore evaluate the signed quantity in Eq.~\eqref{eq:flip_condition} rather than relying only on singular values or subspace angles.

\paragraph{Design implication.} Eq.~(\ref{eq:flip_condition}) separates subspace identification from unlearning. A projection direction can be highly aligned with the target head and still fail to unlearn whenever the target margin remains larger than the induced logit change. Let $a_t=u_1^\top W_t$ and $r=h_t^\top u_1$. The target-logit drop is $ra_t$, so target-head alignment helps only through one term of the condition.
\begin{equation}
  \operatorname{sign}\!\left(m(h_t)-(\delta_t(h_t)+\delta_o(h_t))\right)
  \neq
  \operatorname{sign}\!\left(-|u_1^\top W_t|\right)
  \quad\text{in general}.
\label{eq:alignment_not_enough}
\end{equation}
This is the reason InstantForget does not use the projection direction as the gate or as the correction direction. Detection is calibrated from clean feature statistics, and correction moves flagged features toward a neutral non-target feature rather than subtracting a target-head-aligned component.

\subsection{InstantForget as a Clean-Calibrated Gated Reset}
\label{sec:instantforget}

Projection-style editors use the same estimated direction for detection and correction. InstantForget instead treats these as separate design choices. A clean-calibrated score determines whether intervention is needed, while a neutral reset determines the edited representation. All statistics below are fit from clean splits only. These statistics are $(\mu,\Sigma,M_n,\tau)$. Triggered validation is used at most once beforehand to pick the fixed operating point reported in Sec.~\ref{sec:experiments}, not during deployment calibration.

\subsubsection{Separating Detection from Correction}

Projection directions are learned from the same feature changes that also affect the target classifier head, so using them as a gate risks conflating triggered inputs with clean target-class examples. InstantForget avoids this coupling. A clean-calibrated Mahalanobis score decides whether an input is suspicious, and a neutral reset performs the correction without projecting directly along a target-head-aligned direction.

\subsubsection{Algorithm.}

\paragraph{Clean-calibrated gate.} Given clean fit features $H_{\mathrm{fit}}$, estimate
\begin{equation}
  \mu=\mathbb{E}[h],\qquad
  \Sigma=\operatorname{Cov}(h)+\lambda I.
\end{equation}
The anomaly score is the global Mahalanobis distance
\begin{equation}
  s(h)=(h-\mu)^\top\Sigma^{-1}(h-\mu).
\end{equation}
We calibrate a threshold $\tau$ on a held-out clean validation split so that the clean false-positive rate is a fixed value $q$.

\paragraph{Calibration guarantee.} Let $S=s(f_\ell(X))$ be the clean score random variable and let $\widehat{\tau}_q$ be the empirical $(1-q)$ quantile computed from $n$ held-out clean validation examples. If validation and test clean examples are exchangeable, then the clean false-positive rate is controlled by the quantile error. In particular, by the Dvoretzky-Kiefer-Wolfowitz inequality~\citep{massart1990tight}, with probability at least $1-\delta$,
\begin{equation}
  \Pr\{S>\widehat{\tau}_q\}
  \leq q + \sqrt{\frac{\log(2/\delta)}{2n}} .
\label{eq:fpr_bound}
\end{equation}
Thus the gate has an explicit clean-side control knob. Increasing the calibration split tightens the realized FPR around the chosen target $q$. This guarantee does not say that triggered inputs will be detected. That remains an empirical property of the trigger distribution. It does ensure that the detector is not tuned by inspecting triggered examples.

\paragraph{Neutral non-target reset.} Let
\begin{equation}
  M_n = \mathbb{E}_{(x,y)\in D_c,\ y\neq t}[f_\ell(x)]
\end{equation}
be the mean clean non-target feature. At inference time,
\begin{equation}
\Phi(h)=
\begin{cases}
(1-\beta)h+\beta M_n, & s(h)>\tau,\\
h, & s(h)\leq\tau .
\end{cases}
\end{equation}
The final prediction is $g(\Phi(f_\ell(x)))$.

\paragraph{Deployment cost.} Unlearning is a one-time fit. Extracting $n_{\mathrm{fit}}$ clean features costs $n_{\mathrm{fit}}$ forward passes. Estimating $(\mu,\Sigma^{-1},M_n,\tau)$ costs $O(n_{\mathrm{fit}}d^2+d^3)$ for the $d$-dimensional penultimate feature, where $d=512$ for ResNet-18. No gradient computation or parameter update is used. At inference, the gate adds one quadratic form $O(d^2)$ per input and the reset adds one vector blend. The model weights are untouched, so the edit is trivially reversible and can be deployed or withdrawn per input stream.

\begin{algorithm}[h]
\caption{InstantForget}
\label{alg:instantforget}
\begin{algorithmic}[1]
\REQUIRE Frozen model $g\circ f_\ell$, clean fit split $D_{\mathrm{fit}}$, clean validation split $D_{\mathrm{val}}$, target $t$, FPR target $q$
\STATE $H_{\mathrm{fit}}\gets \{f_\ell(x):(x,y)\in D_{\mathrm{fit}}\}$
\STATE Estimate $\mu,\Sigma^{-1}$ from $H_{\mathrm{fit}}$
\STATE $M_n\gets \operatorname{mean}\{f_\ell(x):(x,y)\in D_{\mathrm{fit}}, y\neq t\}$
\STATE $\tau\gets$ $(1-q)$ quantile of $s(f_\ell(x))$ on $D_{\mathrm{val}}$
\FOR{test input $x$}
\STATE $h\gets f_\ell(x)$
\IF{$(h-\mu)^\top\Sigma^{-1}(h-\mu)>\tau$}
\STATE $h\gets (1-\beta)h+\beta M_n$
\ENDIF
\STATE return $g(h)$
\ENDFOR
\end{algorithmic}
\end{algorithm}

\section{Experiments}
\label{sec:experiments}

\subsection{Experimental Setup}

\paragraph{Datasets.} We use four datasets: CIFAR-10~\citep{krizhevsky2009cifar}, a Brain Tumor MRI dataset~\citep{nickparvar2021braintumor}, the COVID-19 Radiography database~\citep{chowdhury2020covidradiography,rahman2021covidenhancement}, and ModelNet10~\citep{wu2015shapenets}. CIFAR-10 with ResNet-18 is the primary benchmark, evaluated under four attacks that span the common trigger types, namely BadNets (localized patch), Blended (blended), WaNet (geometric warp), and SIG (sinusoidal), with clean calibration and clean testing drawn from disjoint halves of the CIFAR-10 test split. The other three datasets form a cross-domain stress test that reuses the same deployment-calibration protocol: Brain MRI and COVID-19 Radiography use BadNets and Blended image triggers, while ModelNet10 uses point patch and point blend triggers.

\paragraph{Settings.} All main experiments use a 10\% poison ratio, target label 0 unless specified, and three random seeds. Our BadNets trigger is a $5{\times}5$ top-left patch on CIFAR-10 rather than BackdoorBench's default $3{\times}3$. All reported post-training baselines are run on the same poisoned checkpoints, so the comparison is internal to this attack suite. InstantForget uses one fixed operating point unless otherwise stated. We set $n_{\mathrm{fit}}=2000$, clean FPR target $q=0.10$, and reset strength $\beta=1.0$. This point is selected once from a clean-fit and reset grid over $n_{\mathrm{fit}}\in\{500,1000,2000,5000\}$, $q\in\{0.01,0.02,0.05,0.10\}$, and $\beta\in\{0.25,0.5,0.75,1.0\}$ using a guarded validation score that penalizes ASR, clean-accuracy loss, and target-class collapse on a held-out triggered validation split. That grid search is a one-time design choice on the main ResNet-18 checkpoints. At deployment, InstantForget still estimates $(\mu,\Sigma,M_n)$ and calibrates $\tau$ from clean data only. We therefore separate \emph{operating-point selection}, which uses triggered validation in the reported configuration, from \emph{deployment calibration}, which is clean-only.
For the oracle projection audit (Sec.~\ref{sec:audit}), we estimate the trigger subspace $U$ from 5000 paired clean/triggered training samples per trigger and seed, sweep rank $k\in\{1,2,4,8,16,32,64,128,256\}$ and strength $\alpha\in\{0.1,0.3,0.5,0.7,0.9,1.0\}$, and score each operating point by $J=\mathrm{ASR}+\max(\Delta\mathrm{Acc},0)$, where lower is better.

\paragraph{Baselines.} We compare against three families.
\textbf{Post-training repair}: four BackdoorBench defenses with complete coverage on our checkpoints, namely Fine-Pruning, I-BAU, NAD, and NPD~\citep{wu2022backdoorbench}. They modify model parameters, whereas InstantForget edits only inference-time representations of a frozen model. Additional BackdoorBench methods are cited in related work but omitted from the main comparison after failed ResNet-18 integration in our harness; the appendix records why ANP, RNP, CLP, NC, FT, FT-SAM, SAU, and FST were abandoned rather than repaired.
\textbf{Update-free closed-form erasure}: the SVD-of-$D$ projection from the audit, LEACE~\citep{belrose2023neurips-leace}, and a DAMP-style prototype-residual head edit~\citep{hatami2026classunlearningdepthawareremoval}, all receiving oracle paired trigger features.
\textbf{Inference-time input detection}: STRIP~\citep{gao2019strip} and SCALE-UP~\citep{guo2023scaleup}, compared against our gate under one matched calibration protocol. 
We avoid claiming superiority over all backdoor defenses; the post-training rows are limited to the four complete BackdoorBench baselines above.

\paragraph{Metrics.} We report attack success rate (ASR) on triggered non-target test images~\citep{gu2019badnets,wu2022backdoorbench}, clean accuracy, and target-class clean accuracy. For the gate and the input-detection comparison we adopt the standard detection protocol~\citep{gao2019strip,guo2023scaleup}: the threshold-free area under the ROC curve (AUROC), the realized clean test false-positive rate (FPR), and the triggered true-positive rate at the calibrated $10\%$-FPR operating point (TPR@$0.10$, reported as trigger TPR in the main tables). Target-class clean accuracy is treated as a primary metric because backdoor unlearning is selective forgetting. A method that lowers ASR by collapsing the target class has not removed the trigger behavior in a useful way.

\subsection{Auditing Oracle Projection}
\label{sec:audit}

\subsubsection{Oracle gap.}

Oracle SVD-of-$D$ projection removes the backdoor only for BadNets and leaves WaNet, Blended, and SIG near their original ASR, even with paired trigger features and a full rank and strength sweep.

\begin{table}[h]
\centering
\caption{Oracle SVD-of-$D$ audit on CIFAR-10 ResNet-18. Even with paired triggered features and a full hyperparameter sweep, projection succeeds mainly on BadNets.}
\label{tab:audit}
\begin{tabular}{lcccccc}
\toprule
\textbf{Attack} & \textbf{$\epsilon_{k^\star}$} & \textbf{$k_{\mathrm{opt}}$} & \textbf{$\alpha_{\mathrm{opt}}$} & \textbf{ASR$_{\mathrm{opt}}$} & \textbf{$\Delta$Acc$_{\mathrm{opt}}$} & \textbf{$J_{\min}$} \\
\midrule
BadNets & 0.0096\std{0.0004} & 1 & 1.0 & \best{0.139} & 0.019 & \best{0.157\std{0.078}} \\
WaNet   & 0.0139\std{0.0008} & 1 & 1.0 & \second{0.683} & 0.023 & \second{0.706\std{0.053}} \\
Blended & 0.0138\std{0.0001} & 1 & 1.0 & 0.888 & 0.020 & 0.908\std{0.061} \\
SIG     & 0.1640\std{0.2080}   & 1 & 1.0 & 0.941 & 0.008 & 0.950\std{0.017} \\
\bottomrule
\end{tabular}
\end{table}

A large unlearning gap is observed (see Table~\ref{tab:audit}). BadNets reaches ASR 0.139 at the best operating point. WaNet, Blended, and SIG remain at 0.683, 0.888, and 0.941 ASR. All attacks choose the same optimal editor, $k=1,\alpha=1$, so the gap is not caused by hyperparameter mismatch.

The spectral compactness estimate also fails as a stable predictor. For 11 of 12 trigger/seed cells, $\epsilon_{k^\star}$ lies between 0.009 and 0.017. This would suggest that all triggers are compact. The exception is one SIG seed where the knee detector jumps from $k^\star=9$ to $k^\star=1$, which produces the large standard deviation reported in Table~\ref{tab:audit}. Consequently, a single-seed correlation between spectral diffuseness and $J_{\min}$ of 0.976 collapses to 0.275 over the full 12-cell audit. This is a useful negative result. The spectral diagnostic that motivates projection is itself too brittle to explain projection success.

\subsubsection{What governs the gap.}

The gap is governed by a logit-triplet condition at the classifier head, not by spectral compactness, spatial localization, or subspace alignment. We rule out the natural explanations first. Early spatial localization separates BadNets from distributed triggers, but it predicts only a binary patch-vs-distributed split. The observed difficulty is graded. WaNet is easier than Blended and SIG under projection. The trigger subspace is also not misaligned with the model's decision direction. Across triggers, the top SVD direction captures roughly 93 to 97\% of the target-head direction, and the mean principal angle between the trigger subspace and the task subspace is small.

\begin{table}[h]
\centering
\caption{Diagnostics for projection-based unlearning on CIFAR-10 ResNet-18 (four triggers, three seeds). \emph{Spatial localization}: early-layer sparsity separates BadNets from the distributed triggers but cannot explain the graded difficulty among WaNet, Blended, and SIG. \emph{Subspace alignment}: the top SVD direction is almost parallel to the target-class head for every attack, so the failure is not a missed direction. \emph{Logit-triplet}: the signed gap $m-(\delta_t+\delta_o)$ tracks flip rate; a negative gap means projection is likely to flip triggered predictions off the target. Across the 12 trigger/seed cells the gap correlates with flip rate at Pearson $-0.982$ (single-seed $-0.995$), whereas spectral diffuseness $\epsilon_{k^\star}$ drops from a single-seed $+0.976$ to $+0.275$, confirming that the gap is the stable predictor.}
\label{tab:diagnostics}
\resizebox{\linewidth}{!}{
\begin{tabular}{lcccccccc}
\toprule
 & \multicolumn{4}{c}{\textbf{Spatial localization}} & \multicolumn{2}{c}{\textbf{Subspace alignment}} & \multicolumn{2}{c}{\textbf{Logit-triplet}} \\
\cmidrule(lr){2-5}\cmidrule(lr){6-7}\cmidrule(lr){8-9}
\textbf{Attack} & \textbf{R.shift L1} & \textbf{Spars. L1} & \textbf{R.shift L4} & \textbf{Spars. L4} & \textbf{Angle $\rho(U,T)$} & \textbf{$|\langle u_1, W_t/\|W_t\|\rangle|$} & \textbf{Gap $m-(\delta_t+\delta_o)$} & \textbf{Flip rate} \\
\midrule
BadNets & 0.204 & 0.939 & 1.379 & 0.330 & $7.5^\circ$ & 0.960 & -0.667\std{0.170} & 0.858\std{0.079} \\
WaNet   & 0.833 & 0.007 & 1.091 & 0.000 & $3.5^\circ$ & 0.967 & +0.108\std{0.105} & 0.325\std{0.054} \\
Blended & 0.796 & 0.000 & 1.629 & 0.000 & $3.1^\circ$ & 0.961 & +0.444\std{0.137} & 0.118\std{0.063} \\
SIG     & 0.333 & 2.1e-5 & 1.813 & 0.000 & $4.4^\circ$ & 0.928 & +0.754\std{0.088} & 0.060\std{0.016} \\
\bottomrule
\end{tabular}}
\end{table}

The spatial hypothesis is tested in the left block of Table~\ref{tab:diagnostics}. BadNets is localized in early layers, while WaNet, Blended, and SIG are spatially global almost immediately. This explains why BadNets is different, but it does not explain why WaNet is substantially more defendable than Blended and SIG under the same projection.

The alignment result is intentionally unfavorable to our critique. If projection failed merely because SVD missed the relevant direction, low alignment for WaNet, Blended, or SIG would be expected (middle block of Table~\ref{tab:diagnostics}). The opposite is observed. SVD finds a direction almost parallel to the target-class head for every attack. Therefore the remaining explanation must involve how the linear head responds after the projection, not whether the direction was found.

\begin{wrapfigure}{r}{0.46\textwidth}
\centering
\vspace{-\baselineskip}
\includegraphics[width=0.44\textwidth]{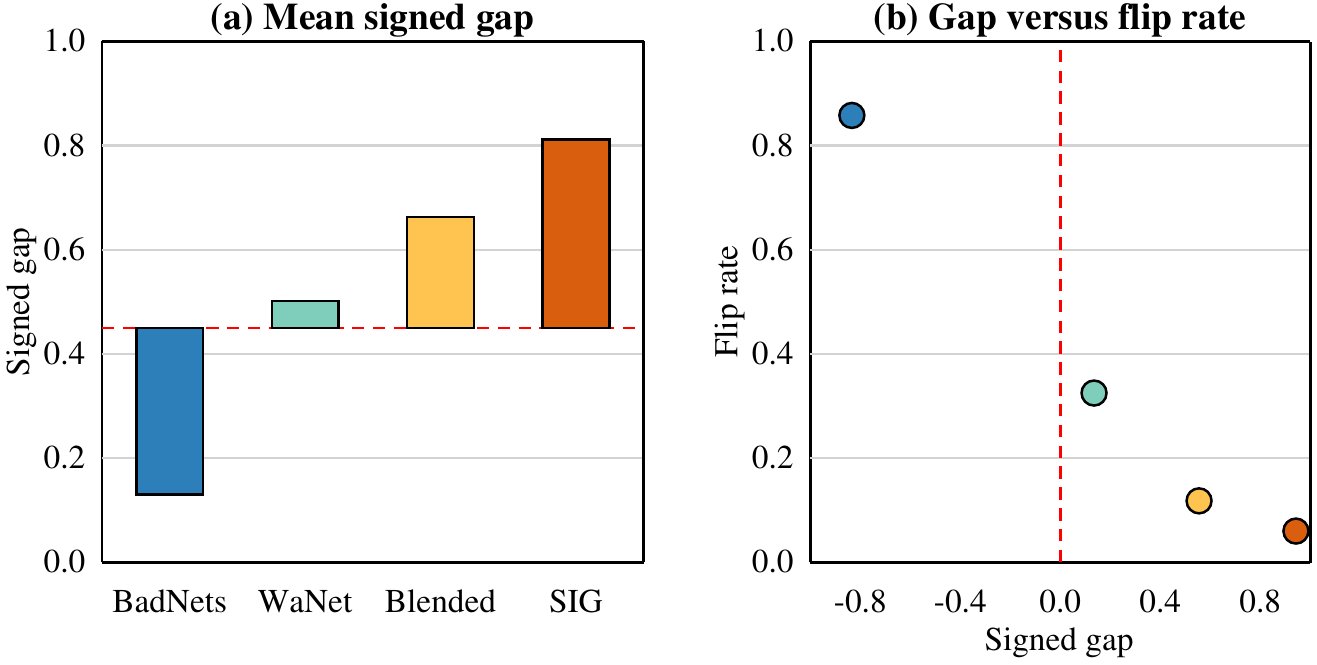}
\caption{Logit-triplet gap mechanism. The signed gap $m-(\delta_t+\delta_o)$ separates projection success from failure and predicts flip rate across trigger families.}
\label{fig:triplet}
\vspace{-\baselineskip}
\vspace{-10pt}
\end{wrapfigure}

We then evaluate the signed quantity in Eq.~(\ref{eq:flip_condition}) on the same projection audit cells. This tests whether the algebraic flip condition, with no fitted predictor, explains which trigger families projection actually changes after the classifier head. The right block of Table~\ref{tab:diagnostics} reports the gap and the resulting flip rate. Across 12 trigger/seed cells, the gap has Pearson correlation $-0.982$ with flip rate. By contrast, the spectral diffuseness diagnostic drops from a single-seed correlation of $0.976$ to $0.275$ across three seeds. SVD does find the target-logit pathway, yet projection fails because the target margin remains too large relative to the combined logit change.

\paragraph{Diagnostic use.} The triplet can be used as a cheap defendability diagnostic for projection-based editing. Given one paired batch, the defender extracts $H_c,H_t$, computes the top right singular vector $u_1$ of $H_t-H_c$, and measures the average values of $m,\delta_t,\delta_o$ on triggered samples. A positive gap predicts that SVD projection will leave many triggered samples classified as the target. A negative gap predicts that projection will often flip them away from the target. The diagnostic is not a final deployment defense because it requires triggered samples, but it is a useful audit tool. It tells us when the projection family should be abandoned rather than tuned harder.

\subsection{Comparison with Post-Training Repair}

\begin{wrapfigure}{r}{0.6\textwidth}
\centering
\vspace{-10pt}
\includegraphics[width=1.0\linewidth]{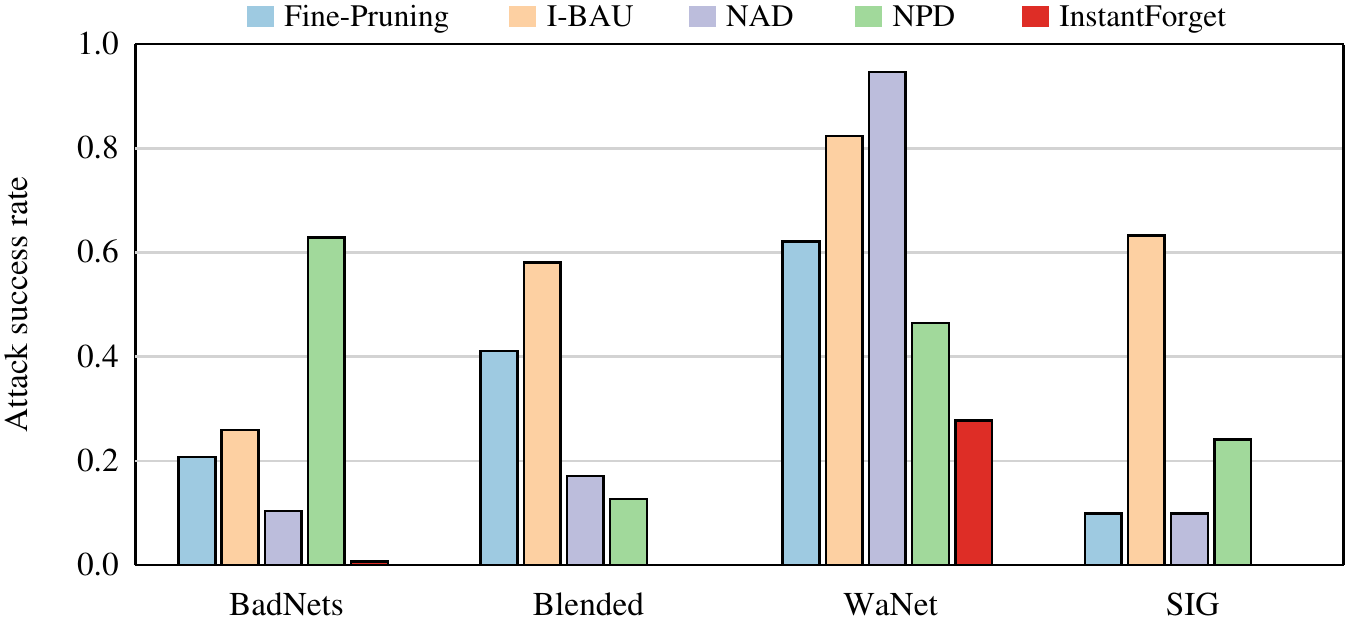}
\caption{Comparison with BackdoorBench baselines on CIFAR-10 ResNet-18. Bars show mean ASR over three seeds. Lower is better. Exact per-trigger values are reported separately (see Table~\ref{tab:comparison_full}).}
\label{fig:comparison_asr}
\vspace{-10pt}
\end{wrapfigure}

\begin{wrapfigure}{r}{0.4\textwidth}
\vspace{-70pt}
\centering
\includegraphics[width=1.0\linewidth]{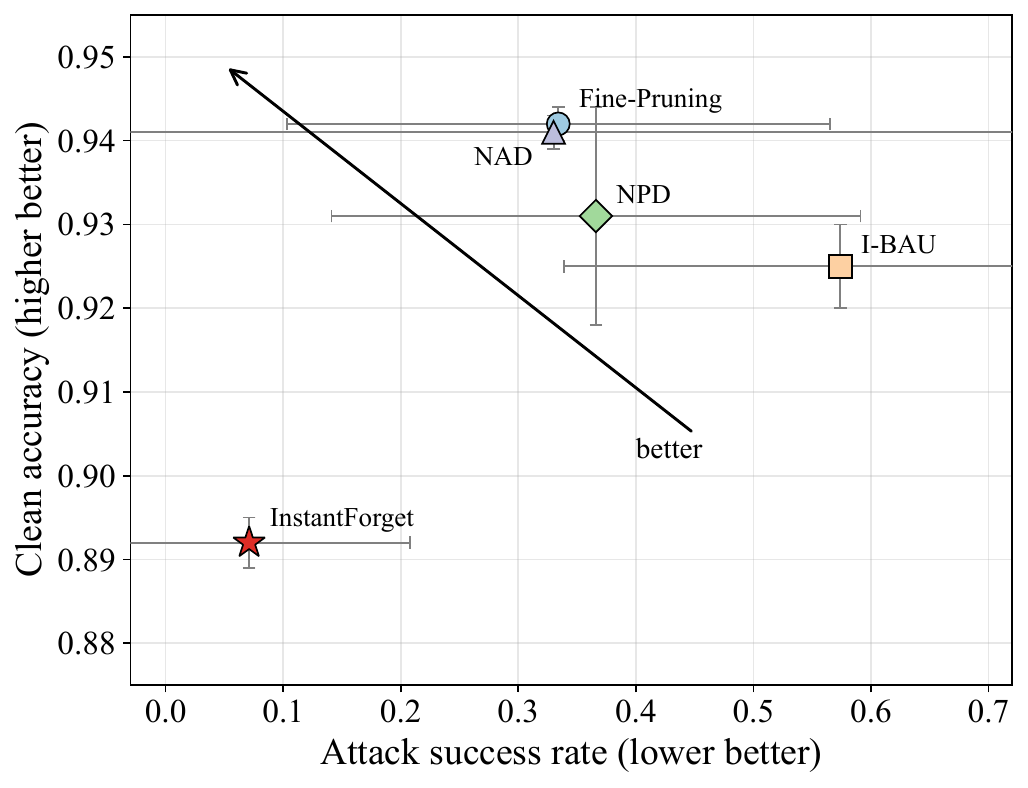}
\caption{ASR and clean-accuracy trade-off on CIFAR-10 ResNet-18, averaged first over three seeds per trigger and then over four triggers. Bars in each direction show the standard deviation across trigger families. Up and to the left is better. InstantForget attains the lowest ASR (0.071) in an update-free regime, trading a few points of clean accuracy against the parameter-updating BackdoorBench defenses. Exact values are given in Table~\ref{tab:comparison_summary}.}
\label{fig:tradeoff}
\vspace{-20pt}
\end{wrapfigure}

InstantForget is compared with four complete BackdoorBench baselines under the same model, poison ratio, and three random seeds (see Figs.~\ref{fig:comparison_asr},~\ref{fig:tradeoff} and Table~\ref{tab:comparison_full}). No post-training baseline handles all trigger families. NAD is strong on BadNets and SIG but fails on WaNet. NPD is strongest among the post-training baselines on Blended and WaNet. InstantForget uses a frozen model and clean deployment calibration, yet the lowest ASR is obtained on every trigger family among these four baselines. On WaNet, InstantForget reaches 0.277 ASR and NPD reaches 0.465. This ASR gain comes with a clean-accuracy cost relative to the post-training defenses and to clean retraining (see Table~\ref{tab:comparison_full}). The main claim is an ASR and utility trade-off in an update-free access regime rather than dominance over retraining-based or parameter-updating defenses. WaNet remains the hardest case, which is consistent with the geometric-trigger failure boundary studied later.

Context for the utility gap is given by the clean-retraining upper bound in Table~\ref{tab:comparison_full}. Retraining from clean data is the strongest upper bound, but it requires clean training data and a new training run. InstantForget is not meant to dominate this oracle. Its value is that it works without retraining or parameter updates.

\subsection{Anatomy of the Gated Reset}

\subsubsection{Closed-form erasure baselines.}
\label{sec:erasure_baselines}

If the projection audit identified merely a weakness of our own SVD editor, a better closed-form eraser should close the gap. We therefore evaluate two standard concept-erasure editors in the same harness, at the same operating point, and with strictly more information than InstantForget receives. Both are fit on oracle paired clean/triggered features. LEACE~\citep{belrose2023neurips-leace} applies the minimal affine edit that removes the linearly encoded triggered-vs-clean concept from the penultimate features. The DAMP-style baseline adapts concurrent class-unlearning surgery~\citep{hatami2026classunlearningdepthawareremoval} to our single-stage setting. The trigger direction is the residual of the triggered prototype after projection onto the span of clean class prototypes. The head weight is right-projected by $W(I-\alpha qq^\top)$, and $\alpha$ is set from held-out probe separability as in the original method.

\begin{wrapfigure}{r}{0.45\textwidth}
\centering
\vspace{-50pt}
\includegraphics[width=1.0\linewidth]{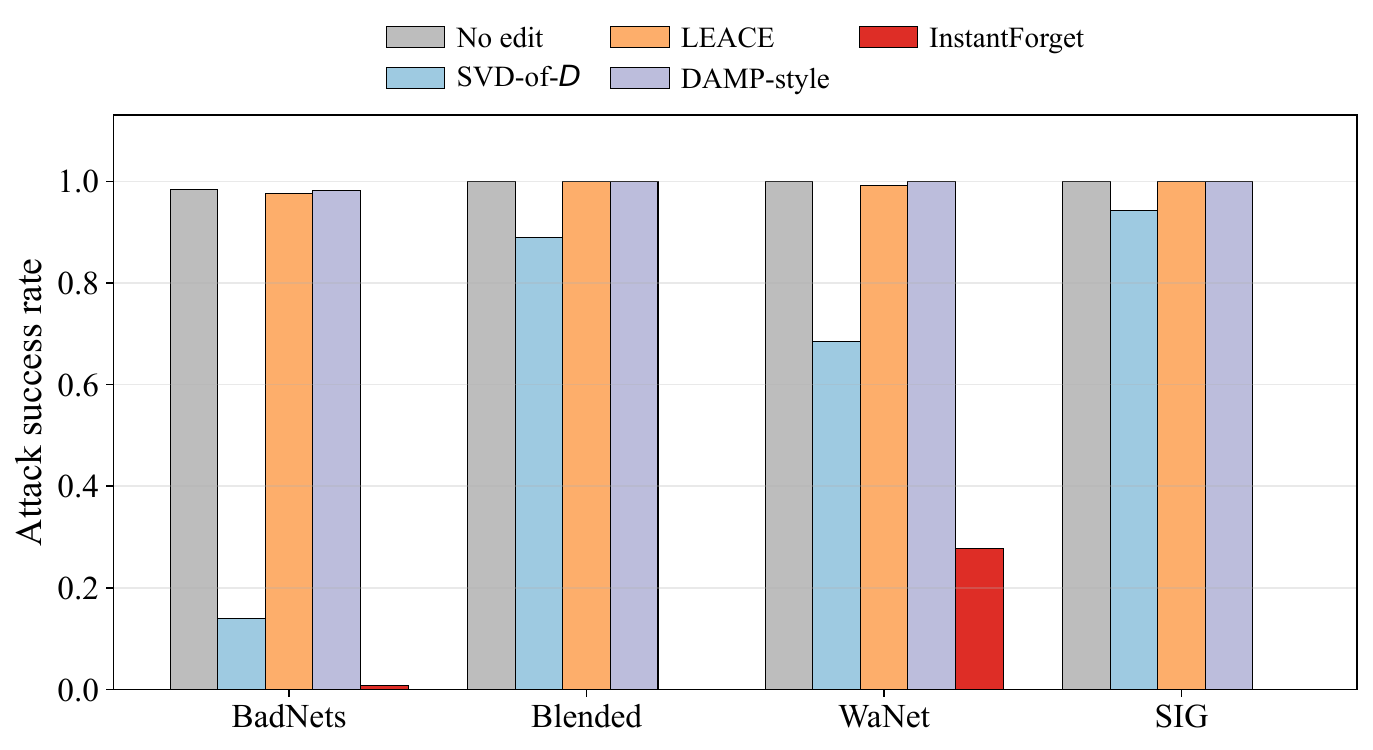}
\caption{Closed-form erasure baselines on CIFAR-10 ResNet-18, mean over three seeds (lower is better). The erasure editors (SVD-of-$D$, LEACE, DAMP-style) receive oracle paired trigger features, yet erasing the linear trigger direction barely changes ASR while preserving target accuracy (LEACE 0.996, DAMP-style 0.969), so the failure is a property of erasure geometry, not of one editor. InstantForget instead reaches average ASR 0.071 at 0.847 target accuracy. Exact values in Table~\ref{tab:erasure_baselines}.}
\label{fig:erasure_asr}
\vspace{-20pt}
\end{wrapfigure}

The result is sharper than the audit alone (see Fig.~\ref{fig:erasure_asr}). LEACE removes the triggered-vs-clean concept direction exactly as designed and preserves utility almost perfectly. Its target accuracy is 0.996, yet average ASR stays at 0.992. The minimal edit that delinearizes the concept does not move triggered inputs off the target prediction, because the erased direction carries the clean target class as much as the trigger. The DAMP-style head edit is similarly inert here. Its probe-separability scaling is calibrated for class unlearning, where the forget class should lose its representation, not for backdoor unlearning, where the target class must survive. This is the constructive reading of the collinearity finding reported earlier (see the subspace-alignment block of Table~\ref{tab:diagnostics}). When the trigger direction and the target-class head are nearly parallel, any linear eraser faces the same dilemma. It can remove too little to change predictions or too much to keep the target class. InstantForget sidesteps the dilemma by not erasing a direction at all.

\subsubsection{Detection quality of the gate.}
\label{sec:detection_comparison}

InstantForget's gate is an inference-time per-input detector, so we also compare it against the standard input-level detection family under one protocol. Thresholds are calibrated on held-out clean validation data at 10\% FPR and evaluated on disjoint clean and triggered test halves. STRIP~\citep{gao2019strip} superimposes clean images and scores prediction entropy. SCALE-UP~\citep{guo2023scaleup} measures scaled-prediction consistency in a black-box manner.

\begin{wrapfigure}{r}{0.4\textwidth}
\centering
\vspace{-15pt}
\includegraphics[width=1.0\linewidth]{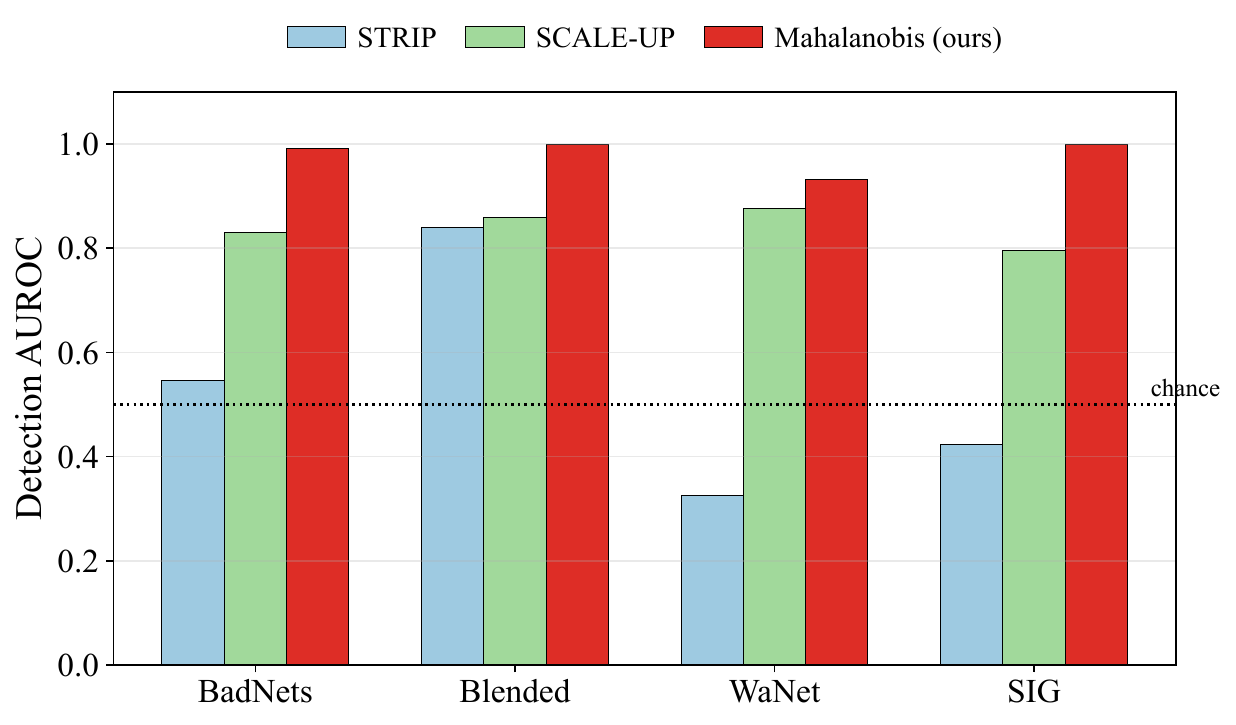}
\caption{Detection AUROC per trigger under matched clean-side calibration (10\% FPR), mean over three seeds. AUROC is threshold-free and is the fair comparison for SCALE-UP, whose discrete consistency score saturates at the 90\% clean quantile and fires on nothing at the calibrated threshold (TPR@0.10 of 0.000). At the calibrated threshold the Mahalanobis gate reaches 0.926 average TPR@0.10 versus 0.214 for STRIP. The clean-calibrated Mahalanobis gate dominates on every trigger family (average AUROC 0.981 vs 0.840 for SCALE-UP and 0.534 for STRIP). Exact values are given in Table~\ref{tab:detection}.}
\label{fig:detection}
\vspace{-20pt}
\end{wrapfigure}

The clean-calibrated Mahalanobis gate dominates on every trigger family (see Fig.~\ref{fig:detection}). The weakest cell is the now-familiar WaNet dip, with AUROC 0.932 and TPR 0.722. STRIP degrades sharply on WaNet and SIG, whose triggers do not survive superposition. SCALE-UP is more uniform but pays for its discrete score with an unusable calibrated operating point. This comparison supports the design decision to build the gate from clean feature statistics rather than from input-space perturbation heuristics.

\subsubsection{Component ablations.}

The central design choice is to separate detection from correction. Projection-only editing receives strictly more information than InstantForget, namely paired triggered features and a per-attack rank/strength sweep. Even so, at its best operating point it leaves average ASR at 0.663 and still fails on Blended, WaNet, and SIG (Table~\ref{tab:audit}), because suppressing the target-head-aligned projection direction is not the same as selectively unlearning the trigger behavior. InstantForget instead uses a clean-calibrated gate to decide when to intervene and a neutral non-target reset to decide where flagged features should move, cutting average ASR to 0.071 at 0.847 target-class accuracy (Table~\ref{tab:component_ablation}). The gain is largest for Blended and SIG, where projection almost never flips the target prediction but the clean-calibrated gate detects the triggered features reliably. WaNet remains partially resistant, which matches the method's stated limitation rather than being hidden by the average.

The detector alone does not change ASR (see Table~\ref{tab:component_ablation}), so unlearning requires an edit after detection. However, the edit cannot be arbitrary. Zero reset and target-mean reset preserve target-class accuracy but leave ASR near one. Resetting every input to the non-target mean drives ASR to zero only by collapsing clean accuracy to random-guess level. The useful operating point is the selective gate paired with the non-target neutral mean. This confirms that the method's gain comes from the combination of clean-calibrated detection and a correction direction that moves flagged features away from the target pathway.

\begin{table}[h]
\centering
\caption{Clean-calibrated component ablation over 12 cells. The cells cover four triggers and three seeds. Cells show mean with standard deviation. Lower ASR is better. Higher accuracy is better. The detector-only row reports the calibrated gate without editing features.}
\label{tab:component_ablation}
\footnotesize
\begin{tabular}{lccccc}
\toprule
\textbf{Variant} & \textbf{ASR} & \textbf{Clean Acc} & \textbf{Target Acc} & \textbf{Test FPR} & \textbf{Trigger TPR} \\
\midrule
No edit & 0.995\std{0.008} & \best{0.954\std{0.002}} & 0.969\std{0.007} & 0.095\std{0.007} & 0.926\std{0.137} \\
Detector only & 0.995\std{0.008} & \best{0.954\std{0.002}} & 0.969\std{0.007} & 0.095\std{0.007} & 0.926\std{0.137} \\
Gate + zero reset & 0.998\std{0.004} & 0.890\std{0.006} & \best{0.984\std{0.006}} & 0.095\std{0.007} & 0.926\std{0.137} \\
Gate + clean mean & 0.831\std{0.388} & 0.890\std{0.006} & 0.958\std{0.061} & 0.095\std{0.007} & 0.926\std{0.137} \\
Gate + target mean & 0.998\std{0.004} & 0.890\std{0.006} & \best{0.984\std{0.006}} & 0.095\std{0.007} & 0.926\std{0.137} \\
Reset all to non-target mean & \best{0.000\std{0.000}} & 0.098\std{0.000} & 0.000\std{0.000} & 1.000\std{0.000} & \best{1.000\std{0.000}} \\
Gate + non-target mean & \second{0.071\std{0.138}} & \second{0.892\std{0.006}} & 0.847\std{0.018} & 0.095\std{0.007} & 0.926\std{0.137} \\
\bottomrule
\end{tabular}
\end{table}

\subsection{Generalization and Failure Boundaries}

\subsubsection{Cross-domain stress.}
\label{sec:cross_domain}

We next test whether the clean-calibrated gate and neutral reset are specific to CIFAR-10. Results are reported on two medical-image datasets and one 3D shape dataset (see Table~\ref{tab:cross_domain}). Brain MRI and COVID-19 Radiography use BadNets and Blended image triggers. ModelNet10 uses point patch and point blend triggers with a PointNet-style classifier.

\begin{table}[h]
\centering
\caption{Cross-domain deployment-calibration stress test. Cells show mean with standard deviation over three seeds. Lower ASR is better. Higher accuracy is better.}
\label{tab:cross_domain}
\footnotesize
\setlength{\tabcolsep}{4pt}
\begin{tabular}{llccccc}
\toprule
\textbf{Dataset} & \textbf{Trigger} & \textbf{ASR} & \textbf{Clean Acc} & \textbf{Target Acc} & \textbf{Test FPR} & \textbf{Trigger TPR} \\
\midrule
Brain MRI & BadNets & \best{0.000\std{0.000}} & 0.869\std{0.017} & 0.792\std{0.014} & 0.083\std{0.010} & \best{1.000\std{0.000}} \\
Brain MRI & Blended & \best{0.000\std{0.000}} & 0.863\std{0.006} & 0.800\std{0.025} & 0.094\std{0.011} & 0.983\std{0.000} \\
COVID Radiography & BadNets & \best{0.000\std{0.000}} & 0.882\std{0.004} & 0.743\std{0.078} & 0.095\std{0.012} & \best{1.000\std{0.000}} \\
COVID Radiography & Blended & \best{0.000\std{0.000}} & 0.876\std{0.010} & 0.795\std{0.044} & 0.098\std{0.003} & \best{1.000\std{0.000}} \\
ModelNet10 & Point patch & \best{0.000\std{0.000}} & 0.835\std{0.008} & 0.560\std{0.040} & 0.066\std{0.010} & \best{1.000\std{0.000}} \\
ModelNet10 & Point blend & 0.514\std{0.052} & 0.811\std{0.012} & 0.627\std{0.046} & 0.067\std{0.017} & 0.171\std{0.042} \\
\bottomrule
\end{tabular}
\end{table}

The medical-image stress tests are strong positives. Average ASR is 0.000 on both Brain MRI and COVID-19 Radiography, with clean accuracy around 0.87 to 0.88. ModelNet10 is mixed. The point patch trigger is removed, but point blend remains high-ASR at 0.514. This result is useful precisely because it is not uniformly favorable. The clean-calibrated feature-distance gate transfers to new domains, but global 3D point shifts can lie too close to the clean feature manifold to be reliably reset.

\subsubsection{Backbone generalization.}
\label{sec:backbone}

\begin{wrapfigure}{r}{0.6\textwidth}
\centering
\vspace{-57pt}
\includegraphics[width=1.0\linewidth]{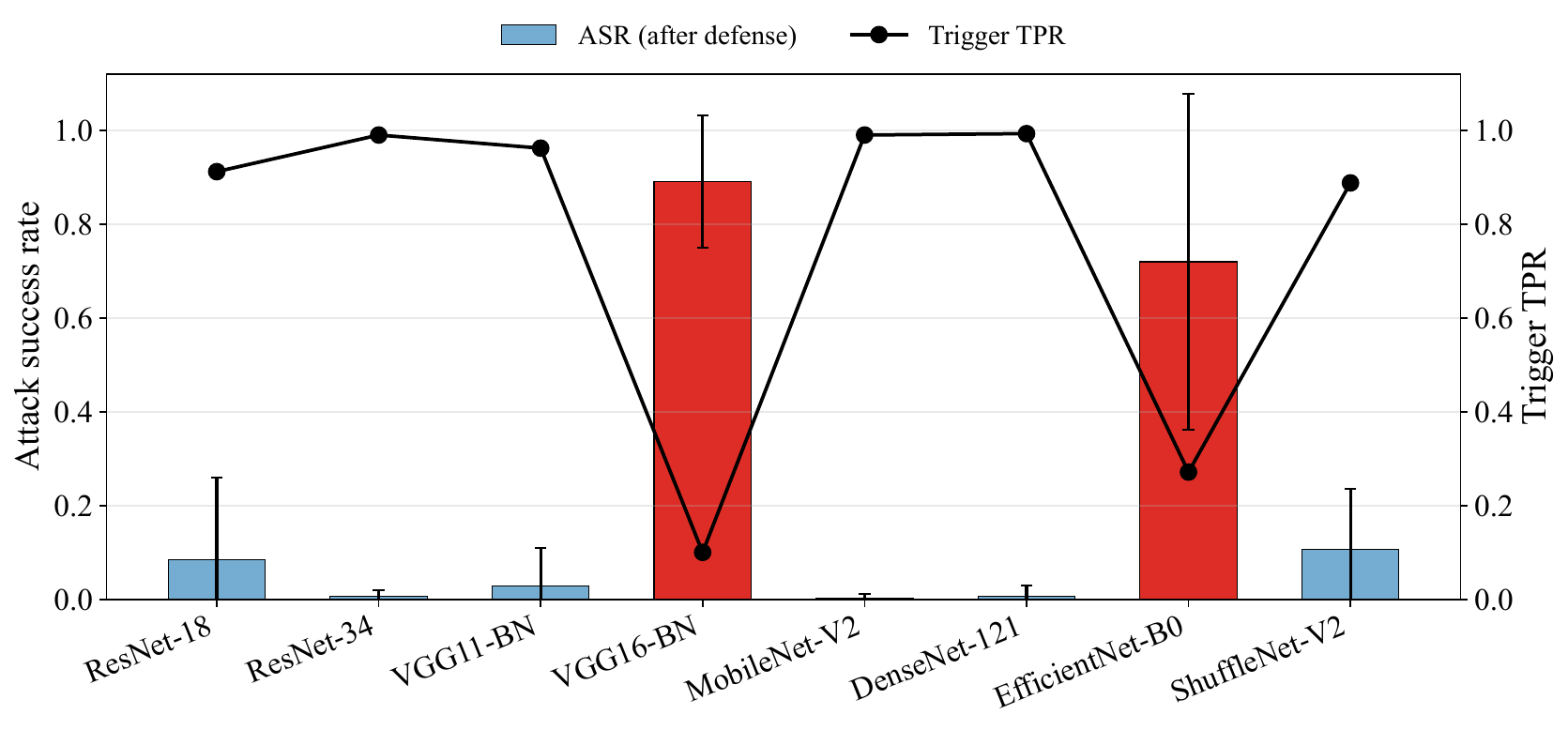}
\caption{Backbone generalization at the fixed operating point. Bars show post-defense ASR (red: failures with $\mathrm{ASR}>0.3$), the line trigger TPR. The two failures (VGG16-BN, EfficientNet-B0) are detection failures, where trigger TPR collapses. Exact values in Table~\ref{tab:backbone}.}
\label{fig:backbone}
\vspace{-20pt}
\end{wrapfigure}

The mechanism makes no ResNet-specific assumption, so we retrain the full poisoned benchmark on eight backbones spanning residual, VGG-style, mobile, and densely connected designs. The identical fixed operating point is applied to each model's penultimate features. These checkpoints are freshly retrained in a separate batch from the main-table ResNet-18 models (see Fig.~\ref{fig:tradeoff}). The ResNet-18 row in the backbone study is therefore an independent sanity check rather than a duplicate of the main result (see Fig.~\ref{fig:backbone}). Its average ASR is 0.086, while the main result is 0.071. Results are aggregated over the four trigger families and three seeds per backbone.

Six of eight backbones transfer well. ResNet-34, DenseNet-121, and MobileNet-V2 reach average ASR below 0.01 with the same hyperparameters chosen on ResNet-18. Even WaNet drops below 0.03 ASR on ResNet-34 and DenseNet-121, which is better than the ResNet-18 main setting. The two failures are informative. On EfficientNet-B0 and VGG16-BN, the trigger TPR collapses to 0.272 and 0.101 while the realized clean FPR stays at the calibrated 0.10. The failure is in detection, not in the reset. Triggered features on these backbones are not separable from clean features by a single global Mahalanobis score at the penultimate layer. This mirrors the WaNet finding at the architecture level. Detectability is a joint property of the trigger and the feature geometry, not a guarantee of the method. We report both failures as scope boundaries rather than tuning targets, since per-backbone layer selection or detector ensembles would change the fixed-operating-point protocol that the rest of the paper holds constant. Per-trigger backbone results are reported separately (see Table~\ref{tab:app_backbone_per_trigger}). The failing backbones fail uniformly across trigger families, not on a single outlier.

\subsubsection{Target labels.}

We train additional poisoned models with target labels 1 and 3 and evaluate the same fixed operating point. Target 1 closely matches the main setting. Average ASR is 0.052, clean accuracy is 0.891, and target accuracy is 0.937. Target 3 exposes a stronger utility trade-off. Average ASR remains low at 0.082 and clean accuracy is 0.885, but target-class accuracy drops to 0.726. InstantForget therefore generalizes in ASR suppression across target labels, but target-class utility depends on target geometry.

\begin{wraptable}{r}{0.6\textwidth}
\centering
\vspace{-20pt}
\caption{Target-label stress test by attack. Results are reported as mean with standard deviation over three rounds.}
\label{tab:target_by_attack}
\footnotesize
\setlength{\tabcolsep}{3pt}
\begin{tabular}{llccc}
\toprule
\textbf{Target} & \textbf{Attack} & \textbf{ASR} & \textbf{Clean Acc} & \textbf{Target Acc} \\
\midrule
1 & BadNets & 0.007\std{0.001} & 0.892\std{0.002} & \best{0.945\std{0.006}} \\
1 & Blended & 0.001\std{0.002} & \best{0.894\std{0.004}} & 0.941\std{0.013} \\
1 & WaNet   & 0.201\std{0.154} & 0.887\std{0.005} & 0.935\std{0.004} \\
1 & SIG     & \best{0.000\std{0.000}} & \second{0.892\std{0.002}} & 0.927\std{0.010} \\
\midrule
3 & BadNets & 0.059\std{0.020} & 0.884\std{0.005} & \best{0.744\std{0.004}} \\
3 & Blended & \best{0.000\std{0.000}} & \best{0.889\std{0.006}} & 0.722\std{0.014} \\
3 & WaNet   & 0.269\std{0.045} & 0.879\std{0.004} & 0.716\std{0.008} \\
3 & SIG     & \best{0.000\std{0.000}} & \second{0.887\std{0.004}} & \second{0.721\std{0.013}} \\
\bottomrule
\end{tabular}
\vspace{-30pt}
\end{wraptable}

The target-label experiment guards against a misleading interpretation of InstantForget. The method does not simply exploit target label 0. ASR suppression transfers to targets 1 and 3. However, the cost is target dependent. Target 3 loses much more target-class accuracy. This suggests that the utility cost is geometric rather than a poorly chosen hyperparameter.

\subsubsection{Adaptive stress.}
\label{sec:adaptive_stress}

InstantForget assumes a non-adaptive training-time attacker. The attacker implants a backdoor but does not explicitly optimize against the deployed gate-and-reset rule. We test this boundary by training poisoned models with an additional feature-compactness penalty that encourages triggered features to remain near the clean feature center while retaining the target prediction. Under this adaptive stress, InstantForget fails to lower ASR (see Table~\ref{tab:adaptive_stress}). The average defended ASR remains 0.996. This failure is not a detection miss: the realized trigger TPR stays high, so many triggered inputs are flagged, but the neutral reset no longer moves them off the attacker-chosen target. This result is not part of the standard comparison because it changes the threat model, but it matters for scope calibration. InstantForget is a post-hoc clean-calibrated unlearning method, not a defense against an attacker who trains the poisoned model to make its target mapping invariant to the reset.

\begin{table}[h]
\centering
\caption{Adaptive feature-compactness stress test. The attacker adds a compactness loss during poisoned training while preserving the target prediction. Results are reported as mean with standard deviation over three rounds.}
\footnotesize
\label{tab:adaptive_stress}
\footnotesize
\setlength{\tabcolsep}{4pt}
\begin{tabular}{lcccccc}
\toprule
\textbf{Attack} & \textbf{No-defense ASR} & \textbf{InstantForget ASR} & \textbf{Clean Acc} & \textbf{Target Acc} & \textbf{Test FPR} & \textbf{Trigger TPR} \\
\midrule
BadNets & 0.984\std{0.001} & 0.985\std{0.001} & 0.881\std{0.002} & 0.974\std{0.005} & 0.100\std{0.000} & 0.974\std{0.000} \\
Blended & 1.000\std{0.000} & 1.000\std{0.000} & 0.889\std{0.002} & 0.977\std{0.008} & 0.100\std{0.000} & 0.909\std{0.087} \\
WaNet   & 0.999\std{0.000} & 0.999\std{0.000} & 0.877\std{0.002} & 0.964\std{0.006} & 0.100\std{0.000} & 0.993\std{0.001} \\
SIG     & 1.000\std{0.000} & 1.000\std{0.000} & 0.885\std{0.003} & 0.974\std{0.003} & 0.100\std{0.000} & 1.000\std{0.000} \\
\midrule
Average & 0.996\std{0.007} & 0.996\std{0.007} & 0.883\std{0.005} & 0.972\std{0.007} & 0.100\std{0.000} & 0.969\std{0.053} \\
\bottomrule
\end{tabular}
\end{table}

\section{Discussion}
\label{sec:discussion}

The main result is not that a Mahalanobis detector universally solves backdoor unlearning. The result is more specific. Projection failure can be predicted by a logit-level quantity. The failure extends to closed-form erasers such as LEACE and DAMP-style head edits even with oracle trigger access. Once detection is separated from target-head erasure, a simple clean-calibrated reset provides a stronger ASR and utility trade-off than the tested baselines in this benchmark.

\paragraph{When does update-free unlearning work?} The positive and negative results in this paper fit a two-part criterion. Triggered inputs must be detectable under the clean feature distribution of the deployed backbone. The neutral reset must also move flagged features across the target decision boundary. This criterion is orthogonal to the one that governs erasure. LEACE and the projection family ask whether the trigger has a removable linear direction. It does, but that direction is target-entangled and removing it does not flip predictions. The gate instead asks whether triggered features are distributionally separable. This holds for additive triggers on most backbones, weakens for the geometric WaNet warp, and fails for EfficientNet-B0 and VGG16-BN feature spaces and for ModelNet10 point blend. The adaptive stress test exposes the second part of the criterion. Even when many triggered inputs are flagged, an attacker can train the target mapping to survive the reset. We consider making both quantities checkable before deployment, e.g.\ from realized TPR on suspected inputs and post-reset target margins, a more useful direction for update-free defense than pursuing a universal eraser.

\paragraph{Limitations.} InstantForget should not be read as exact unlearning or as a free lunch. It reduces clean accuracy and target-class clean accuracy, with target label 3 exposing a larger utility cost. WaNet on CIFAR-10 and point blend on ModelNet10 remain difficult, suggesting that geometric or global feature shifts can evade a substantial fraction of the clean-calibrated gate. The backbone study sharpens this boundary. On EfficientNet-B0 and VGG16-BN the gate's trigger TPR collapses at the fixed operating point, so the method inherits the detectability of triggered features in a given feature geometry rather than guaranteeing it. The adaptive stress test gives the sharpest boundary. When the attacker explicitly trains the poisoned model so that triggered features remain target-predictive after a neutral reset, InstantForget does not reduce ASR even though many examples are flagged. The method also assumes access to clean calibration data. The fixed operating point reported in the main tables was selected once with held-out triggered validation. On the comparison side, the post-training table is scoped to the four BackdoorBench defenses with complete logs; the remaining methods were omitted after integration failures rather than reported as incomplete rows. Since the closest update-free erasure baselines, LEACE and a DAMP-style editor, are already evaluated in Sec.~\ref{sec:erasure_baselines}, this scoping does not affect the paper's central claims.

\section{Conclusion}
\label{sec:conclusion}

We audited projection-based update-free backdoor unlearning under oracle conditions and found a large, predictable failure gap. The failure is not explained by spectral compactness alone, spatial locality, or subspace alignment. It is governed by a logit-triplet gap that determines whether projection changes the decision, and it persists for closed-form concept erasers given strictly more access than our method. This diagnosis motivates InstantForget, a clean-calibrated gated reset that decouples detection from correction. Without retraining or parameter updates, the method gives the strongest ASR reduction among the tested non-adaptive defenses in our benchmark. It also outperforms STRIP and SCALE-UP as a detector and transfers across most backbones at a fixed operating point. The broader lesson is that update-free repair is possible, but only when detection, correction, target-class utility, and adaptive robustness are evaluated separately.

\clearpage
\bibliographystyle{plainnat}
\bibliography{main}

\clearpage
\appendix
\setcounter{figure}{0}
\setcounter{table}{0}
\renewcommand{\thefigure}{A\arabic{figure}}
\renewcommand{\thetable}{A\arabic{table}}

\section*{Appendix}
\input{appendix_projection_evidence}

\begin{table}[h]
\centering
\caption{Comparison with four complete BackdoorBench baselines, plus a clean-retraining upper bound. Cells show ASR and clean accuracy, reported as mean with standard deviation over three seeds. BackdoorBench logs report overall clean accuracy rather than target-class accuracy; InstantForget reports both in the ablation tables. The bottom row is a clean-retraining upper bound: it removes the backdoor almost completely (target-class accuracy 0.964) but requires training a new model from clean data. Lower ASR is better.}
\label{tab:comparison_full}
\footnotesize
\resizebox{\linewidth}{!}{
\begin{tabular}{lcccc}
\toprule
\textbf{Defense} & \textbf{BadNets} & \textbf{Blended} & \textbf{WaNet} & \textbf{SIG} \\
\midrule
Fine-Pruning & \second{0.207\std{0.025}} (0.943\std{0.002}) & 0.411\std{0.084} (0.940\std{0.003}) & 0.621\std{0.077} (0.944\std{0.002}) & \second{0.099\std{0.005}} (0.941\std{0.003}) \\
I-BAU        & 0.259\std{0.154} (0.925\std{0.013}) & 0.581\std{0.175} (0.921\std{0.007}) & 0.824\std{0.118} (0.932\std{0.006}) & 0.633\std{0.418} (0.924\std{0.002}) \\
NAD          & 0.104\std{0.001} (0.941\std{0.009}) & 0.171\std{0.062} (0.939\std{0.008}) & 0.947\std{0.065} (0.943\std{0.007}) & \second{0.099\std{0.004}} (0.940\std{0.007}) \\
NPD          & 0.629\std{0.220} (0.935\std{0.002}) & \second{0.127\std{0.002}} (0.939\std{0.002}) & \second{0.465\std{0.464}} (0.937\std{0.001}) & 0.241\std{0.241} (0.912\std{0.026}) \\
InstantForget & \best{0.007\std{0.005}} (0.894\std{0.006}) & \best{0.000\std{0.000}} (0.893\std{0.007}) & \best{0.277\std{0.141}} (0.893\std{0.005}) & \best{0.000\std{0.000}} (0.887\std{0.005}) \\
\midrule
\textit{Clean retraining}$^{\dagger}$ & 0.007\std{0.001} (0.953\std{0.003}) & 0.001\std{0.000} (0.953\std{0.003}) & 0.070\std{0.010} (0.953\std{0.003}) & 0.002\std{0.000} (0.953\std{0.003}) \\
\bottomrule
\end{tabular}
}
\\[2pt]
{\scriptsize $^{\dagger}$Upper bound: trains a new model from clean data (average ASR 0.020, target-class accuracy 0.964).}
\end{table}

\end{document}

%% file: math_commands.tex
\usepackage{amsmath,amsfonts,bm}

\def\eqref#1{equation~\ref{#1}}

\def\1{\bm{1}}

\DeclareMathAlphabet{\mathsfit}{\encodingdefault}{\sfdefault}{m}{sl}
\SetMathAlphabet{\mathsfit}{bold}{\encodingdefault}{\sfdefault}{bx}{n}



%% file: appendix_projection_evidence.tex
\appendix \setcounter{section}{0} \renewcommand{\thesection}{\Alph{section}} \renewcommand{\theHsection}{appendix.\Alph{section}}

\section{Additional Evidence}
\label{app:local_evidence}

This appendix provides additional diagnostics supporting the projection audit in the main paper. The results are included to clarify why projection is a useful diagnostic but an unsafe unlearning mechanism.

\subsection{Projection Sweep Details}

The SVD-of-$D$ projection audit in the main paper is backed by 12 local sweep cells (see Table~\ref{tab:app_t2_projection}). Each cell contains four triggers, three seeds, and 54 projection settings. The sweep reinforces the same conclusion as the main audit. BadNets is the only trigger family for which projection reaches a low-ASR operating point. Blended, WaNet, and SIG remain high-ASR even when the rank and projection strength are selected by the guarded objective.

\begin{table}[h]
\centering
\caption{Projection sweep summary. Results use $\kappa=1$ and are reported as mean with standard deviation over three seeds.}
\label{tab:app_t2_projection}
\footnotesize
\begin{tabular}{lcccc}
\toprule
\textbf{Trigger} & \textbf{ASR at Opt.} & \textbf{$\Delta$Clean Acc} & \textbf{$J_{\kappa=1}$} & \textbf{$\epsilon_{k^*}$} \\
\midrule
BadNets & \best{0.139\std{0.098}} & 0.019\std{0.004} & \best{0.157\std{0.095}} & 0.010 \\
Blended & 0.888\std{0.073} & 0.020\std{0.004} & 0.908\std{0.074} & 0.014 \\
WaNet   & 0.683\std{0.066} & 0.023\std{0.003} & 0.706\std{0.065} & 0.014 \\
SIG     & 0.941\std{0.020} & \best{0.008\std{0.001}} & 0.950\std{0.020} & 0.164 \\
\bottomrule
\end{tabular}
\end{table}

\subsection{Causal Feature-Shift Diagnostics}

The causal-tracing experiment measures how much the trigger changes intermediate feature blocks and how spatially concentrated that change is (see Table~\ref{tab:app_shift}). BadNets is highly localized early in the network, while Blended and SIG are spatially diffuse from the first measured block. WaNet is only weakly localized at early blocks and becomes diffuse in deeper blocks. This supports the paper's distinction between localized patch triggers and distributed/geometric triggers.

\begin{table}[h]
\centering
\caption{Feature-shift diagnostics, seed 0. Early sparsity is measured in an early residual block. Last-block sparsity and cosine distance are measured at the final residual block.}
\label{tab:app_shift}
\footnotesize
\begin{tabular}{lcccc}
\toprule
\textbf{Trigger} & \textbf{Early Spatial Sparsity} & \textbf{Last Spatial Sparsity} & \textbf{Max Rel. Shift} & \textbf{Last Cos. Dist.} \\
\midrule
BadNets & \best{0.939} & 0.001 & 1.379 & 0.552 \\
Blended & 0.000 & 0.000 & 1.629 & 0.683 \\
WaNet   & 0.007 & 0.000 & 1.091 & 0.630 \\
SIG     & 0.000 & 0.000 & \best{1.813} & \best{0.697} \\
\bottomrule
\end{tabular}
\end{table}

\subsection{Principal-Angle and Target-Head Alignment}

The principal-angle experiment compares the estimated trigger subspace with a target-class task subspace (see Table~\ref{tab:app_principal}). All trigger families show small principal angles and high alignment with the target-class classifier head. This is the local evidence behind the paper's claim that projection directions are not safe correction directions. They are entangled with the target-class decision pathway, so suppressing them can also suppress useful target-class information.

\begin{table}[h]
\centering
\caption{Principal-angle diagnostics, seed 0. Angles are in degrees. $g$ denotes the target-class classifier-head direction.}
\label{tab:app_principal}
\footnotesize
\begin{tabular}{lccccc}
\toprule
\textbf{Trigger} & \textbf{Min Angle} & \textbf{Mean Angle} & \textbf{Top-1 $g$ Align.} & \textbf{Frac. $g$ in $U_9$} & \textbf{Frac. $g$ in Task} \\
\midrule
BadNets & 4.45 & 7.51 & 0.960 & 0.923 & 0.936 \\
Blended & \best{1.00} & \best{3.13} & 0.961 & 0.924 & \best{0.959} \\
WaNet   & 1.73 & 3.49 & \best{0.967} & \best{0.940} & 0.953 \\
SIG     & 1.73 & 4.43 & 0.928 & 0.865 & 0.952 \\
\bottomrule
\end{tabular}
\end{table}

\subsection{Operating-Point Robustness}

We sweep four clean-fit sizes, four clean-FPR targets, and four reset strengths.
\begin{equation}
\begin{aligned}
n_{\mathrm{fit}}&\in\{500,1000,2000,5000\},\\
q&\in\{0.01,0.02,0.05,0.10\},\\
\beta&\in\{0.25,0.5,0.75,1.0\}.
\end{aligned}
\end{equation}
The fixed main-table configuration is also the best single global configuration under the held-out triggered-validation guarded score
\begin{equation}
\mathrm{ASR}+\max(0,0.90-\mathrm{Acc})+\max(0,0.85-\mathrm{Acc}_t),
\end{equation}
computed across all 12 target-0 cells. Thus the main result is not selected separately per attack or per seed.

\begin{table}[h]
\centering
\caption{Best guarded operating point aggregated over three seeds. Per-cell tuning is summarized as mean with standard deviation.}
\label{tab:operating_points}
\footnotesize
\begin{tabular}{lcccc}
\toprule
\textbf{Attack} & \textbf{Test FPR} & \textbf{Trigger TPR} & \textbf{ASR} & \textbf{Clean Acc} \\
\midrule
BadNets & 0.095\std{0.007} & 0.985\std{0.005} & \second{0.007\std{0.005}} & 0.894\std{0.007} \\
Blended & \best{0.034\std{0.024}} & \best{1.000\std{0.000}} & \best{0.000\std{0.000}} & 0.928\std{0.018} \\
WaNet   & 0.092\std{0.006} & 0.722\std{0.141} & 0.277\std{0.142} & 0.893\std{0.005} \\
SIG     & \second{0.026\std{0.018}} & \best{1.000\std{0.000}} & \best{0.000\std{0.000}} & \best{0.936\std{0.010}} \\
\bottomrule
\end{tabular}
\end{table}

Several attacks prefer a lower clean FPR than the fixed operating point, especially Blended and SIG (see Table~\ref{tab:operating_points}). We nevertheless use the single global configuration in the main comparison because it is the best guarded point on average and avoids per-attack tuning. WaNet again determines the trade-off. Reducing its ASR requires a high-FPR, full-strength reset.

\subsection{Backbone Generalization Details}

Per-trigger defended ASR for the backbone study is reported in Table~\ref{tab:app_backbone_per_trigger}. The two failing backbones fail across all four trigger families rather than on one outlier. This supports the feature-geometry reading. The penultimate features of EfficientNet-B0 and VGG16-BN place triggered inputs inside the clean Mahalanobis envelope regardless of trigger type. Conversely, the strongest backbones suppress even WaNet, the hardest trigger in the ResNet-18 main setting, below 0.03 ASR.

\begin{table}[h]
\centering
\caption{Per-trigger defended ASR by backbone at the fixed operating point, mean with standard deviation over three seeds. Lower is better.}
\label{tab:app_backbone_per_trigger}
\footnotesize
\begin{tabular}{lcccc}
\toprule
\textbf{Backbone} & \textbf{BadNets} & \textbf{Blended} & \textbf{WaNet} & \textbf{SIG} \\
\midrule
ResNet-18 & 0.006\std{0.003} & 0.000\std{0.000} & 0.336\std{0.203} & 0.000\std{0.000} \\
ResNet-34 & 0.005\std{0.002} & 0.000\std{0.000} & 0.024\std{0.023} & 0.000\std{0.000} \\
VGG11-BN & 0.010\std{0.006} & 0.000\std{0.000} & 0.111\std{0.148} & 0.000\std{0.001} \\
VGG16-BN & 0.831\std{0.082} & 0.998\std{0.002} & 0.816\std{0.242} & 0.920\std{0.107} \\
MobileNet-V2 & 0.008\std{0.003} & 0.001\std{0.002} & 0.009\std{0.014} & 0.000\std{0.000} \\
DenseNet-121 & 0.000\std{0.000} & 0.000\std{0.000} & 0.027\std{0.046} & 0.000\std{0.000} \\
EfficientNet-B0 & 0.840\std{0.226} & 0.888\std{0.158} & 0.486\std{0.377} & 0.666\std{0.577} \\
ShuffleNet-V2 & 0.106\std{0.164} & 0.131\std{0.120} & 0.191\std{0.151} & 0.000\std{0.000} \\
\bottomrule
\end{tabular}
\end{table}

\subsection{Tabular Versions of Main-Text Figures}
\label{app:figure_tables}

For completeness, this subsection lists the exact numeric values behind the figures in the main text. Table~\ref{tab:comparison_summary} backs Fig.~\ref{fig:tradeoff}, Table~\ref{tab:erasure_baselines} backs Fig.~\ref{fig:erasure_asr}, Table~\ref{tab:detection} backs Fig.~\ref{fig:detection}, and Table~\ref{tab:backbone} backs Fig.~\ref{fig:backbone}.

\begin{table}[h]
\centering
\caption{Overall ASR and utility comparison on CIFAR-10 ResNet-18 (tabular version of Fig.~\ref{fig:tradeoff}). Values are first averaged over three seeds for each trigger and then averaged over four triggers. The small term reports standard deviation across trigger families. Lower ASR is better. Higher clean accuracy is better.}
\label{tab:comparison_summary}
\footnotesize
\begin{tabular}{lcc}
\toprule
\textbf{Defense} & \textbf{ASR $\downarrow$} & \textbf{Clean Acc $\uparrow$} \\
\midrule
Fine-Pruning & 0.334\std{0.231} & \best{0.942\std{0.002}} \\
I-BAU & 0.574\std{0.235} & 0.925\std{0.005} \\
NAD & \second{0.330\std{0.412}} & \second{0.941\std{0.002}} \\
NPD & 0.366\std{0.225} & 0.931\std{0.013} \\
InstantForget & \best{0.071\std{0.137}} & 0.892\std{0.003} \\
\bottomrule
\end{tabular}
\end{table}

\begin{table}[h]
\centering
\caption{Closed-form erasure baselines on CIFAR-10 ResNet-18, mean over three seeds (tabular version of Fig.~\ref{fig:erasure_asr}). All erasure editors receive oracle paired trigger features. InstantForget uses clean deployment calibration at the fixed operating point. Lower ASR is better.}
\label{tab:erasure_baselines}
\footnotesize
\begin{tabular}{lccccccc}
\toprule
\textbf{Editor} & \textbf{Trigger access} & \textbf{BadNets} & \textbf{Blended} & \textbf{WaNet} & \textbf{SIG} & \textbf{Avg ASR} & \textbf{Avg Target Acc} \\
\midrule
No edit & none & 0.983 & 1.000 & 0.999 & 1.000 & 0.995\std{0.008} & 0.969\std{0.007} \\
SVD-of-$D$ & yes & \second{0.140} & \second{0.890} & \second{0.685} & \second{0.943} & \second{0.665\std{0.337}} & 0.765\std{0.067} \\
LEACE & yes & 0.977 & 1.000 & 0.992 & 1.000 & 0.992\std{0.010} & \best{0.996\std{0.002}} \\
DAMP-style & yes & 0.982 & 1.000 & 0.999 & 1.000 & 0.995\std{0.008} & 0.969\std{0.007} \\
InstantForget & no & \best{0.007} & \best{0.000} & \best{0.277} & \best{0.000} & \best{0.071\std{0.138}} & 0.847\std{0.018} \\
\bottomrule
\end{tabular}
\end{table}

\begin{table}[h]
\centering
\caption{Detection comparison under matched clean-side calibration, mean over three seeds (tabular version of Fig.~\ref{fig:detection}). AUROC is threshold-free. TPR is measured at the calibrated threshold. SCALE-UP's discrete consistency score saturates at the 90\% clean quantile, so its calibrated threshold fires on nothing. AUROC is the fair comparison for it.}
\label{tab:detection}
\footnotesize
\begin{tabular}{lcccccc}
\toprule
\textbf{Detector} & \textbf{BadNets} & \textbf{Blended} & \textbf{WaNet} & \textbf{SIG} & \textbf{Avg AUROC} & \textbf{TPR@0.10} \\
\midrule
STRIP & 0.546 & \second{0.840} & 0.325 & 0.423 & 0.534\std{0.217} & 0.214\std{0.250} \\
SCALE-UP & \second{0.830} & 0.859 & \second{0.876} & \second{0.796} & \second{0.840\std{0.038}} & 0.000\std{0.000} \\
Mahalanobis gate (ours) & \best{0.991} & \best{1.000} & \best{0.932} & \best{1.000} & \best{0.981\std{0.032}} & \best{0.926\std{0.137}} \\
\bottomrule
\end{tabular}
\end{table}

\begin{table}[h]
\centering
\caption{Backbone generalization at the fixed operating point, aggregated over four triggers and three seeds (tabular version of Fig.~\ref{fig:backbone}). Each backbone has 12 cells. Lower ASR is better. Trigger TPR shows that the two failures are detection failures. The gate misses triggered features, so no reset is applied.}
\label{tab:backbone}
\footnotesize
\begin{tabular}{lccccc}
\toprule
\textbf{Backbone} & \textbf{No-defense ASR} & \textbf{ASR} & \textbf{Clean Acc} & \textbf{Target Acc} & \textbf{Trigger TPR} \\
\midrule
ResNet-18 & 0.995\std{0.008} & 0.086\std{0.174} & 0.891\std{0.004} & 0.849\std{0.018} & 0.912\std{0.173} \\
ResNet-34 & 0.996\std{0.006} & \best{0.007\std{0.014}} & \best{0.896\std{0.007}} & 0.879\std{0.018} & 0.990\std{0.014} \\
VGG11-BN & 0.989\std{0.016} & 0.030\std{0.080} & 0.860\std{0.007} & 0.827\std{0.024} & 0.962\std{0.080} \\
VGG16-BN & 0.992\std{0.013} & 0.891\std{0.141} & 0.894\std{0.003} & 0.889\std{0.017} & 0.101\std{0.138} \\
MobileNet-V2 & 0.992\std{0.013} & \best{0.004\std{0.008}} & 0.830\std{0.007} & 0.812\std{0.025} & \best{0.990\std{0.013}} \\
DenseNet-121 & 0.997\std{0.005} & \best{0.007\std{0.023}} & 0.889\std{0.006} & 0.882\std{0.024} & \best{0.993\std{0.023}} \\
EfficientNet-B0 & 0.992\std{0.012} & 0.720\std{0.357} & 0.822\std{0.009} & 0.849\std{0.028} & 0.272\std{0.359} \\
ShuffleNet-V2 & 0.994\std{0.009} & 0.107\std{0.130} & 0.828\std{0.006} & 0.777\std{0.023} & 0.888\std{0.131} \\
\bottomrule
\end{tabular}
\end{table}

\subsection{Omitted BackdoorBench Defenses}
\label{app:extended_baselines}

We initially planned a broader BackdoorBench comparison with ANP, RNP, CLP, NC, FT, FT-SAM, SAU, and FST on the same ResNet-18 checkpoints. Only Fine-Pruning, I-BAU, NAD, and NPD completed all 12 trigger/seed cells and therefore enter the main table. The remaining defenses were not tabulated because their logs were incomplete or not parseable under our harness. RNP ran on only six cells. In those cells, clean accuracy collapsed to roughly 0.10 while BadNets ASR remained at 1.000. ANP failed with a ResNet-18 integration error. CLP, NC, FT, FT-SAM, SAU, and FST produced no parseable final-epoch logs. Since Sec.~\ref{sec:erasure_baselines} already compares InstantForget to LEACE and a DAMP-style editor under a matched protocol, omitting these incomplete BackdoorBench rows does not weaken the paper's update-free erasure critique.

%% file: main.bib
@article{gu2019badnets,
  title={BadNets: Evaluating Backdooring Attacks on Deep Neural Networks},
  author={Gu, Tianyu and Liu, Kang and Dolan-Gavitt, Brendan and Garg, Siddharth},
  journal={IEEE Access},
  volume={7},
  pages={47230--47244},
  year={2019},
  doi={10.1109/ACCESS.2019.2909068}
}

@inproceedings{nguyen2021wanet,
  title={WaNet: Imperceptible Warping-Based Backdoor Attack},
  author={Nguyen, Tuan Anh and Tran, Anh Tuan},
  booktitle={International Conference on Learning Representations},
  year={2021},
  url={https://openreview.net/forum?id=eEn8KT0JOQ}
}

@misc{chen2017targetedbackdoorattacksdeep,
  title={Targeted Backdoor Attacks on Deep Learning Systems Using Data Poisoning},
  author={Chen, Xinyun and Liu, Chang and Li, Bo and Lu, Kimberly and Song, Dawn},
  year={2017},
  eprint={1712.05526},
  archivePrefix={arXiv},
  primaryClass={cs.CR},
  url={https://arxiv.org/abs/1712.05526}
}

@inproceedings{barni2019new,
  title={A New Backdoor Attack in {CNNS} by Training Set Corruption Without Label Poisoning},
  author={Barni, Mauro and Kallas, Kassem and Tondi, Benedetta},
  booktitle={2019 IEEE International Conference on Image Processing (ICIP)},
  pages={101--105},
  year={2019},
  organization={IEEE},
  doi={10.1109/ICIP.2019.8802997}
}

@inproceedings{li2021invisible,
  title={Invisible Backdoor Attack with Sample-Specific Triggers},
  author={Li, Yuezun and Li, Yiming and Wu, Baoyuan and Li, Longkang and He, Ran and Lyu, Siwei},
  booktitle={Proceedings of the IEEE/CVF International Conference on Computer Vision},
  pages={16463--16472},
  year={2021}
}

@inproceedings{liu2018fine,
  title={Fine-Pruning: Defending Against Backdooring Attacks on Deep Neural Networks},
  author={Liu, Kang and Dolan-Gavitt, Brendan and Garg, Siddharth},
  booktitle={International Symposium on Research in Attacks, Intrusions, and Defenses},
  pages={273--294},
  year={2018}
}

@inproceedings{li2021neural,
  title={Neural Attention Distillation: Erasing Backdoor Triggers from Deep Neural Networks},
  author={Li, Yige and Lyu, Xixiang and Koren, Nodens and Lyu, Lingjuan and Li, Bo and Ma, Xingjun},
  booktitle={International Conference on Learning Representations},
  year={2021},
  url={https://openreview.net/forum?id=9l9BDlFHu5}
}

@inproceedings{wu2021adversarial,
  title={Adversarial Neuron Pruning Purifies Backdoored Deep Models},
  author={Wu, Dongxian and Wang, Yisen},
  booktitle={Advances in Neural Information Processing Systems},
  volume={34},
  pages={16913--16925},
  year={2021}
}

@inproceedings{zeng2022adversarial,
  title={Adversarial Unlearning of Backdoors via Implicit Hypergradient},
  author={Zeng, Yi and Chen, Si and Park, Won and Mao, Z. Morley and Jin, Ming and Jia, Ruoxi},
  booktitle={International Conference on Learning Representations},
  year={2022},
  url={https://openreview.net/forum?id=MeeQkFYVbzW}
}

@inproceedings{pang2023backdoor,
  title={Backdoor Cleansing with Unlabeled Data},
  author={Pang, Lu and Sun, Tao and Ling, Haibin and Chen, Chao},
  booktitle={Proceedings of the IEEE/CVF Conference on Computer Vision and Pattern Recognition},
  pages={12218--12227},
  year={2023}
}

@inproceedings{tran2018spectral,
  title={Spectral Signatures in Backdoor Attacks},
  author={Tran, Brandon and Li, Jerry and Madry, Aleksander},
  booktitle={Advances in Neural Information Processing Systems},
  volume={31},
  pages={8000--8010},
  year={2018}
}

@inproceedings{chen2018detectingbackdoorattacksdeep,
  title={Detecting Backdoor Attacks on Deep Neural Networks by Activation Clustering},
  author={Chen, Bryant and Carvalho, Wilka and Baracaldo, Nathalie and Ludwig, Heiko and Edwards, Benjamin and Lee, Taesung and Molloy, Ian and Srivastava, Biplav},
  booktitle={Workshop on Artificial Intelligence Safety 2019 co-located with AAAI},
  year={2019},
  eprint={1811.03728},
  archivePrefix={arXiv},
  url={https://arxiv.org/abs/1811.03728}
}

@inproceedings{wang2019neural,
  title={Neural Cleanse: Identifying and Mitigating Backdoor Attacks in Neural Networks},
  author={Wang, Bolun and Yao, Yuanshun and Shan, Shawn and Li, Huiying and Viswanath, Bimal and Zheng, Haitao and Zhao, Ben Y.},
  booktitle={IEEE Symposium on Security and Privacy},
  pages={707--723},
  year={2019}
}

@inproceedings{gao2019strip,
  title={{STRIP}: A Defence Against Trojan Attacks on Deep Neural Networks},
  author={Gao, Yansong and Xu, Chang and Wang, Derui and Chen, Shiping and Ranasinghe, Damith C. and Nepal, Surya},
  booktitle={Annual Computer Security Applications Conference},
  pages={113--125},
  year={2019}
}

@inproceedings{li2021abl,
  title={Anti-Backdoor Learning: Training Clean Models on Poisoned Data},
  author={Li, Yige and Lyu, Xixiang and Koren, Nodens and Lyu, Lingjuan and Li, Bo and Ma, Xingjun},
  booktitle={Advances in Neural Information Processing Systems},
  volume={34},
  pages={14900--14912},
  year={2021}
}

@inproceedings{wu2022backdoorbench,
  title={BackdoorBench: A Comprehensive Benchmark of Backdoor Learning},
  author={Wu, Baoyuan and Chen, Hongrui and Zhang, Mingda and Zhu, Zihao and Wei, Shaokui and Yuan, Danni and Shen, Chao},
  booktitle={Advances in Neural Information Processing Systems Datasets and Benchmarks Track},
  year={2022},
  url={https://openreview.net/forum?id=31_Uzi-r1Yu}
}

@inproceedings{wei2023shared,
  title={Shared Adversarial Unlearning: Backdoor Mitigation by Unlearning Shared Adversarial Examples},
  author={Wei, Shaokui and Zhang, Mingda and Zha, Hongyuan and Wu, Baoyuan},
  booktitle={Advances in Neural Information Processing Systems},
  year={2023},
  url={https://openreview.net/forum?id=zqOcW3R9rd}
}

@inproceedings{li2023rnp,
  title={Reconstructive Neuron Pruning for Backdoor Defense},
  author={Li, Yige and Lyu, Xixiang and Ma, Xingjun and Koren, Nodens and Lyu, Lingjuan and Li, Bo and Jiang, Yu-Gang},
  booktitle={International Conference on Machine Learning},
  pages={19837--19854},
  year={2023}
}

@inproceedings{zhu2024neurips-breaking,
  title={{Breaking the False Sense of Security in Backdoor Defense Through Re-Activation Attack}},
  author={Zhu, Mingli and Liang, Siyuan and Wu, Baoyuan},
  booktitle={Advances in Neural Information Processing Systems},
  year={2024},
  doi={10.52202/079017-3649},
  url={https://mlanthology.org/neurips/2024/zhu2024neurips-breaking/}
}

@inproceedings{bourtoule2021machine,
  title={Machine Unlearning},
  author={Bourtoule, Lucas and Chandrasekaran, Varun and Choquette-Choo, Christopher A. and Jia, Hengrui and Travers, Adelin and Zhang, Baiwu and Lie, David and Papernot, Nicolas},
  booktitle={2021 IEEE Symposium on Security and Privacy},
  pages={141--159},
  year={2021}
}

@inproceedings{neel2021descent,
  title={Descent-to-Delete: Gradient-Based Methods for Machine Unlearning},
  author={Neel, Seth and Roth, Aaron and Sharifi-Malvajerdi, Saeed},
  booktitle={Proceedings of the 32nd International Conference on Algorithmic Learning Theory},
  series={Proceedings of Machine Learning Research},
  volume={132},
  pages={931--962},
  year={2021}
}

@inproceedings{guo2020certified,
  title={Certified Data Removal from Machine Learning Models},
  author={Guo, Chuan and Goldstein, Tom and Hannun, Awni and van der Maaten, Laurens},
  booktitle={Proceedings of the 37th International Conference on Machine Learning},
  series={Proceedings of Machine Learning Research},
  volume={119},
  pages={3832--3842},
  year={2020}
}

@article{massart1990tight,
  title={The Tight Constant in the Dvoretzky-Kiefer-Wolfowitz Inequality},
  author={Massart, Pascal},
  journal={The Annals of Probability},
  volume={18},
  number={3},
  pages={1269--1283},
  year={1990}
}

@article{xu2023survey,
  title={Machine Unlearning: A Survey},
  author={Xu, Heng and Zhu, Tianqing and Zhang, Lefeng and Zhou, Wanlei and Yu, Philip S.},
  journal={ACM Computing Surveys},
  volume={56},
  number={1},
  pages={1--36},
  year={2023},
  doi={10.1145/3603620}
}

@inproceedings{pawelczyk2025iclr-machine,
  title={{Machine Unlearning Fails to Remove Data Poisoning Attacks}},
  author={Pawelczyk, Martin and Di, Jimmy Z. and Lu, Yiwei and Kamath, Gautam and Sekhari, Ayush and Neel, Seth},
  booktitle={International Conference on Learning Representations},
  year={2025},
  url={https://mlanthology.org/iclr/2025/pawelczyk2025iclr-machine/}
}

@inproceedings{allouah2025iclr-utility,
  title={{The Utility and Complexity of In- and Out-of-Distribution Machine Unlearning}},
  author={Allouah, Youssef and Kazdan, Joshua and Guerraoui, Rachid and Koyejo, Sanmi},
  booktitle={International Conference on Learning Representations},
  year={2025},
  url={https://mlanthology.org/iclr/2025/allouah2025iclr-utility/}
}

@inproceedings{koloskova2025icml-certified,
  title={{Certified Unlearning for Neural Networks}},
  author={Koloskova, Anastasia and Allouah, Youssef and Jha, Animesh and Guerraoui, Rachid and Koyejo, Sanmi},
  booktitle={Proceedings of the 42nd International Conference on Machine Learning},
  pages={31275--31298},
  volume={267},
  year={2025},
  url={https://mlanthology.org/icml/2025/koloskova2025icml-certified/}
}

@inproceedings{cai2025targeted,
  title={{Targeted Unlearning with Single Layer Unlearning Gradient}},
  author={Cai, Zikui and Tan, Yaoteng and Asif, M. Salman},
  booktitle={Proceedings of the 42nd International Conference on Machine Learning},
  pages={6257--6290},
  year={2025},
  url={https://openreview.net/forum?id=6Ofb0cGXb5}
}

@inproceedings{cooper2025neurips-machine,
  title={{Machine Unlearning Doesn't Do What You Think: Lessons for Generative AI Policy and Research}},
  author={Cooper, A. Feder and Choquette-Choo, Christopher A. and Bogen, Miranda and Klyman, Kevin and Jagielski, Matthew and Filippova, Katja and Liu, Ken and Chouldechova, Alexandra and Hayes, Jamie and Huang, Yangsibo and Triantafillou, Eleni and Kairouz, Peter and Mitchell, Nicole Elyse and Mireshghallah, Niloofar and Jacobs, Abigail Z. and Grimmelmann, James and Shmatikov, Vitaly and De Sa, Christopher and Shumailov, Ilia and Terzis, Andreas and Barocas, Solon and Vaughan, Jennifer Wortman and Boyd, Danah and Choi, Yejin and Koyejo, Sanmi and Delgado, Fernando and Liang, Percy and Ho, Daniel E. and Samuelson, Pamela and Brundage, Miles and Bau, David and Neel, Seth and Wallach, Hanna and Cyphert, Amy B. and Lemley, Mark and Papernot, Nicolas and Lee, Katherine},
  booktitle={Advances in Neural Information Processing Systems},
  year={2025},
  url={https://mlanthology.org/neurips/2025/cooper2025neurips-machine/}
}

@misc{su2025burnbackdoorunlearningadversarial,
  title={{BURN}: Backdoor Unlearning via Adversarial Boundary Analysis},
  author={Su, Yanghao and Zhang, Jie and Li, Yiming and Zhang, Tianwei and Guo, Qing and Zhang, Weiming and Yu, Nenghai and Lukas, Nils and Zhou, Wenbo},
  year={2025},
  eprint={2507.10491},
  archivePrefix={arXiv},
  primaryClass={cs.CR},
  url={https://arxiv.org/abs/2507.10491}
}

@misc{jiang2025backdoortokenunlearningexposing,
  title={Backdoor Token Unlearning: Exposing and Defending Backdoors in Pretrained Language Models},
  author={Jiang, Peihai and Lyu, Xixiang and Li, Yige and Ma, Jing},
  year={2025},
  eprint={2501.03272},
  archivePrefix={arXiv},
  primaryClass={cs.CR},
  url={https://arxiv.org/abs/2501.03272}
}

@misc{abad2025soklinedefensebackdoor,
  title={{SoK}: The Last Line of Defense: On Backdoor Defense Evaluation},
  author={Abad, Gorka and Kr{\v{c}}ek, Marina and Koffas, Stefanos and Tajalli, Behrad and Arazzi, Marco and Ria{\~n}o, Roberto and Xu, Xiaoyun and Liu, Zhuoran and Nocera, Antonino and Picek, Stjepan},
  year={2025},
  eprint={2511.13143},
  archivePrefix={arXiv},
  primaryClass={cs.CR},
  url={https://arxiv.org/abs/2511.13143}
}

@misc{lu2025badfubackdoorfederatedlearning,
  title={{BadFU}: Backdoor Federated Learning through Adversarial Machine Unlearning},
  author={Lu, Bingguang and Hu, Hongsheng and Miao, Yuantian and Sohail, Shaleeza and He, Chaoxiang and Wang, Shuo and Chen, Xiao},
  year={2025},
  eprint={2508.15541},
  archivePrefix={arXiv},
  primaryClass={cs.CR},
  url={https://arxiv.org/abs/2508.15541}
}

@misc{alam2025reveilunconstrainedconcealedbackdoor,
  title={{ReVeil}: Unconstrained Concealed Backdoor Attack on Deep Neural Networks using Machine Unlearning},
  author={Alam, Manaar and Lamri, Hithem and Maniatakos, Michail},
  year={2025},
  eprint={2502.11687},
  archivePrefix={arXiv},
  primaryClass={cs.CR},
  url={https://arxiv.org/abs/2502.11687},
  note={Accepted at 62nd Design Automation Conference (DAC)}
}

@inproceedings{belrose2023neurips-leace,
  title={{LEACE}: Perfect Linear Concept Erasure in Closed Form},
  author={Belrose, Nora and Schneider-Joseph, David and Ravfogel, Shauli and Cotterell, Ryan and Raff, Edward and Biderman, Stella},
  booktitle={Advances in Neural Information Processing Systems},
  year={2023},
  url={https://mlanthology.org/neurips/2023/belrose2023neurips-leace/}
}

@inproceedings{hatami2026classunlearningdepthawareremoval,
  title={Class Unlearning via Depth-Aware Removal of Forget-Specific Directions},
  author={Hatami, Arman and Aalishah, Romina and Monosov, Ilya E.},
  booktitle={Proceedings of the IEEE/CVF Conference on Computer Vision and Pattern Recognition Workshops},
  year={2026},
  eprint={2604.15166},
  archivePrefix={arXiv},
  primaryClass={cs.CV},
  url={https://arxiv.org/abs/2604.15166}
}

@inproceedings{guo2023scaleup,
  title={{SCALE-UP}: An Efficient Black-box Input-level Backdoor Detection via Analyzing Scaled Prediction Consistency},
  author={Guo, Junfeng and Li, Yiming and Chen, Xun and Guo, Hanqing and Sun, Lichao and Liu, Cong},
  booktitle={International Conference on Learning Representations},
  year={2023},
  url={https://openreview.net/forum?id=o0LFPcoFKnr}
}

@techreport{krizhevsky2009cifar,
  title={Learning Multiple Layers of Features from Tiny Images},
  author={Krizhevsky, Alex},
  year={2009},
  institution={University of Toronto},
  url={https://www.cs.toronto.edu/~kriz/learning-features-2009-TR.pdf}
}

@inproceedings{wu2015shapenets,
  title={{3D ShapeNets}: A Deep Representation for Volumetric Shapes},
  author={Wu, Zhirong and Song, Shuran and Khosla, Aditya and Yu, Fisher and Zhang, Linguang and Tang, Xiaoou and Xiao, Jianxiong},
  booktitle={IEEE Conference on Computer Vision and Pattern Recognition},
  pages={1912--1920},
  year={2015},
  doi={10.1109/CVPR.2015.7298801}
}

@misc{nickparvar2021braintumor,
  title={Brain Tumor {MRI} Dataset},
  author={Nickparvar, Masoud},
  year={2021},
  publisher={Kaggle},
  doi={10.34740/KAGGLE/DSV/2645886},
  url={https://www.kaggle.com/datasets/masoudnickparvar/brain-tumor-mri-dataset}
}

@article{chowdhury2020covidradiography,
  title={Can {AI} Help in Screening Viral and {COVID-19} Pneumonia?},
  author={Chowdhury, Muhammad E. H. and Rahman, Tawsifur and Khandakar, Amith and Mazhar, Rashid and Kadir, Muhammad Abdul and Mahbub, Zaid Bin and Islam, Khandakar Reajul and Khan, Muhammad Salman and Iqbal, Atif and Al Emadi, Nasser and Reaz, Mamun Bin Ibne and Islam, Mohammad Tariqul},
  journal={IEEE Access},
  volume={8},
  pages={132665--132676},
  year={2020},
  doi={10.1109/ACCESS.2020.3010287}
}

@article{rahman2021covidenhancement,
  title={Exploring the Effect of Image Enhancement Techniques on {COVID-19} Detection Using Chest {X-ray} Images},
  author={Rahman, Tawsifur and Khandakar, Amith and Qiblawey, Yazan and Tahir, Anas and Kiranyaz, Serkan and Abul Kashem, Saad Bin and Islam, Mohammad Tariqul and Al Maadeed, Somaya and Zughaier, Susu M. and Khan, Muhammad Salman and Chowdhury, Muhammad E. H.},
  journal={Computers in Biology and Medicine},
  volume={132},
  pages={104319},
  year={2021},
  doi={10.1016/j.compbiomed.2021.104319}
}

@article{yu2025forgetme,
  title={{ForgetMe}: Benchmarking the Selective Forgetting Capabilities of Generative Models},
  author={Yu, Zhenyu and Idris, Mohd Yamani Idna and Wang, Pei and Xia, Yuelong and Xiang, Yong},
  journal={Engineering Applications of Artificial Intelligence},
  volume={161},
  pages={112087},
  year={2025},
  publisher={Elsevier},
  doi={10.1016/j.engappai.2025.112087}
}
